\pdfoutput=1
\documentclass[11pt]{article}

\usepackage{acl}
\usepackage{times}
\usepackage{latexsym}

\usepackage[T1]{fontenc}
\usepackage[utf8]{inputenc}

\usepackage{microtype}

\usepackage{inconsolata}
\usepackage{makecell}
\usepackage{graphicx}
\usepackage{booktabs}
\usepackage{geometry}
\usepackage{tabularx}
\usepackage{subcaption}

\title{Emergent Abilities in Reduced-Scale Generative Language Models}

\author{Sherin Muckatira, Vijeta Deshpande, Vladislav Lialin, Anna Rumshisky \\ 
University of Massachusetts Lowell\\ 
\texttt{\{sherinbojappa\_muckatira,  vijeta\_deshpande\}@student.uml.edu} \\
\texttt{\{vlialin, arum\}@cs.uml.edu}  }

\begin{document}
\maketitle
\begin{abstract}

Large language models can solve new tasks without task-specific fine-tuning. This ability, also known as in-context learning (ICL), is considered an emergent ability and is primarily seen in large language models with billions of parameters. This study investigates if such emergent properties are strictly tied to model size or can be demonstrated by smaller models trained on reduced-scale data. To explore this, we simplify pre-training data and pre-train 36 causal language models with parameters varying from 1 million to 165 million parameters.
We show that models trained on this simplified pre-training data demonstrate enhanced zero-shot capabilities across various tasks in simplified language, achieving performance comparable to that of pre-trained models six times larger on unrestricted language. 
This suggests that downscaling the language allows zero-shot learning capabilities to emerge in models with limited size.
%This suggests that downscaling the language facilitates the emergence of zero-shot learning capabilities in smaller-sized models.
% We do find that few-shot prompting does not help these models. 
Additionally, we find that these smaller models pre-trained on simplified data demonstrate a power law relationship between the evaluation loss and the three scaling factors: compute, dataset size, and model size.\footnote{Code and simplified pre-training data are available at \href{https://github.com/text-machine-lab/mini_gpt}{github.com/text-machine-lab/mini\_gpt}}

% To test our hypothesis, we pre-trained XXX causal language models and evaluated models on XXX downstream tasks. We find that language simplification improves the in-context learning abilities of smaller language models. Models trained on simple language achieved +XXX, and +XXX scores on XXX- and XXX-shot settings, compared to the model trained on full language. Hence, vocabulary is a critical component of the smaller language models that needs appropriate scaling as we reduce the size of the language model. 
\end{abstract}

\section{Introduction}

% The benefits of the size of language models are evident in the literature, especially when it comes to in-context learning (ICL) abilities \cite{}. 

Recent advancements in deep learning and distributed computing have enabled the pre-training of language models on a massive scale \citep{brown2020language, bubeck2023sparks, touvron2023llama}, significantly changing the way these models are used. Large pre-trained models proved capable of solving various tasks with zero-shot or few-shot learning, eliminating the need for task-specific fine-tuning \citep{brown2020language}. This is referred to as in-context learning, an ability which allows these models to understand and solve new tasks based on the provided context. It is argued that this ability ``emerges'' with a dramatic increase in the size of the model \citep{wei2022emergent}. % not quite sure how this furthers the narrative considering we do use "non linear" metrics like accuracy.
Efforts to transfer emergent abilities to small models include imitation learning, where a large language model like GPT-4 acts as a ``teacher'' to create synthetic datasets with additional instructions and explanations. This synthetic data is then used to train smaller ``student'' models  \citep{alpaca, peng2023instruction, mukherjee2023orca, magister-etal-2023-teaching}. Another approach is distillation where the ``student'' model is trained to mimic the output probabilities of the ``teacher'' model \citep{gu2023minillm, xu2024survey} .

Our work takes a different approach; our goal is to determine whether simplifying the pre-training data itself can unlock emergent language abilities in smaller models.  This idea is supported by our previous work \cite{deshpande2023honey}, which highlighted the effects of language simplification for smaller models when fine-tuning on downstream tasks. Prior work by \citet{eldan2023tinystories} reports a similar trend, though their approach requires the use of larger models to produce the simplified language. We bypass this step and instead rely on naturally-occurring language restricted via vocabulary filtering.

%We instead focus on the  naturally-occurring language, using restricted via vocabulary filtering, which allows us to eliminate the need for large models to distill the language.

%An alternative strategy to improve emergent abilities in smaller language models involves simplifying the pre-training data. \citet{eldan2023tinystories} demonstrate that this simplification can lead to emergence of language capabilities in smaller models. However, they generated simplified training data using larger models.
%\citet{deshpande2023honey} also emphasize the advantages of language simplification for smaller language models. %($\leq 20 mil.$).  
%However, they focused on downstream evaluation after fine-tuning. 
%and did not investigate the ICL properties of the smaller models.

%In recent work conducted by   In addition, human language acquisition literature as well highlights the language understanding benefits of language simplification.

%motivate this a little better by drawing parallels with psychology
%Prior work such as \citet{eldan2023tinystories} have suggested that simplifying the pre-training data can enhance the coherence and performance of smaller models. However, these approaches primarily rely on generating synthetic data from larger models to simplify the pre-training data, which can be costly and challenging to scale up. 

%Extending the available evidence of benefits due to language simplification, 
%Therefore, in this study, we take this investigation one step further. 
In this study, we leverage this approach to determine whether language simplification can unlock ICL abilities in smaller language models. To do so, we pre-train 36 causal language models with sizes varying from 1M to 165M parameters, on both a simplified English dataset and a standard pre-training dataset and conduct zero-shot evaluations on different tasks. Through extensive experimentation, we show that language simplification enables ICL abilities in smaller language models on a level comparable to larger-size models pre-trained on non-simplified English corpora. \newline

\begin{figure*}
    \centering
    \includegraphics[width=\textwidth]{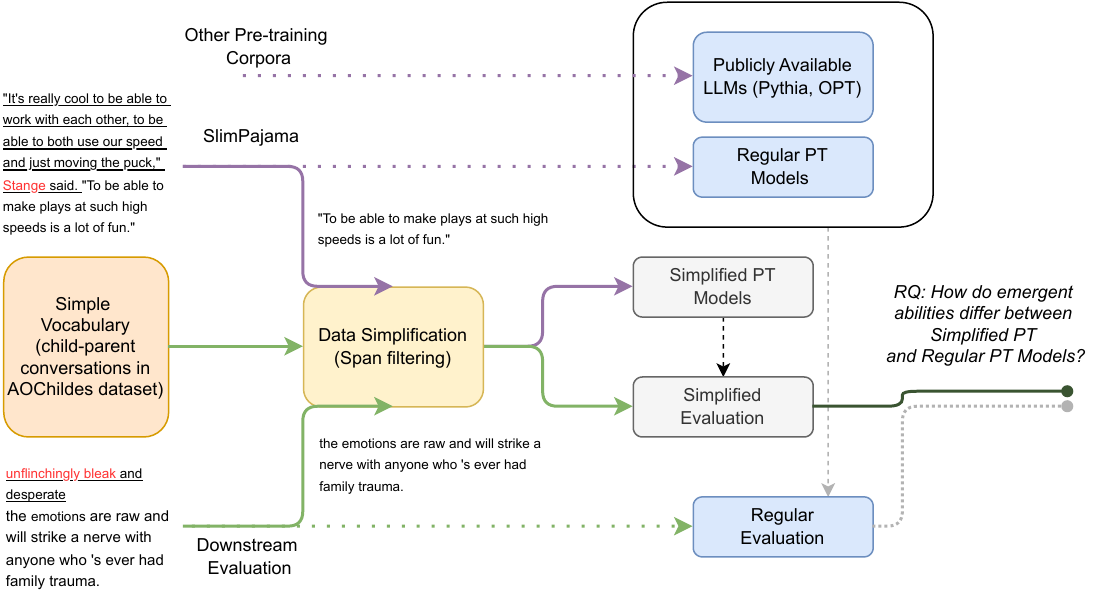}
    \caption{We filter the SlimPajama dataset by selecting spans that contain words from the AO-Childes vocabulary and removing any spans with words not in this vocabulary. We also filter examples in the downstream evaluation dataset based on the occurrence of words in the AO-Childes Vocabulary. The underlined spans are removed by filtering due to the presence of Out of Vocabulary words (Out of Vocabulary words are in red). This simplified dataset is used to pre-train simplified models, whereas regular models are trained on the standard SlimPajama dataset or on other existing pre-training corpora. We then compare whether simplified pre-trained models can perform downstream tasks in simplified language as effectively as standard pre-trained models do in the complete language.}

    \label{fig:method}
\end{figure*}

\noindent{Specifically, our contributions are as follows:}

\begin{itemize}
    \item We demonstrate that downscaling (simplifying) the language enhances zero-shot learning capabilities in smaller-sized models.
    \item We show that small models trained with such simplified data demonstrate a power law relationship between evaluation loss and the three scale factors: FLOPs, Dataset Size, and Model Size.
    \item We release a simplified pre-training corpus obtained by filtering the existing SimPajama dataset \citep{cerebras2023slimpajama}.
    %we obtain a simplified pre-training corpus and we pre-train different-sized small language models (< 165M Parameters).
    % \item We show that the zero-shot performance on grammar-related and other tasks improves due to language simplification. 
\end{itemize}

%In this study, we investigate if small language models, pre-trained on simplified data, can effectively handle tasks they haven't seen before. We simplify standard pre-training data based on child vocabulary, creating a less complex dataset. Our research primarily examines how this simpler pre-training affects the zero-shot abilities of small models. Through experiments with various model sizes, we find that such pre-training improves their performance on grammar and simple tasks. This underscores the significance of dataset complexity in training smaller language models.

% \noindent
% Our code and data will be published on GitHub upon acceptance.

% from Methodology for better placement
% dataset dist

\begin{table}
\centering
\footnotesize
\begin{tabular}{lrr}
\toprule
\textbf{Split} & \textbf{Percentage of} & \textbf{Number of}\\
\textbf{} & \textbf{tokens} & \textbf{tokens (mil)}\\
\midrule
{C4} & {23.86\%} & {5258.73}  \\
{GitHub} & {0.21\%} & {46.10}\\
{Commoncrawl} & {22.12\%} & {4875.09} \\ 
{StackExchange} & {1.33\%} & {293.06} \\ 
{Wikpedia} & {0.08\%} & {18.49}\\
{ArXiv} & {0.53\%}  & {117.66}\\
{Books} & {51.86\%} & {11429.27}\\ 
\midrule
{Total} & {100\%} & {22038.41} \\
\bottomrule

\end{tabular}
\caption{Data source distribution for the simplified pre-training dataset derived from SlimPajama.}
\label{tab:dataset_simple_22B}
\end{table}
% \input{tables/table_dataset_dist}

% From results for better table placement
% \input{tables/table_training_curves}
\section{Related Work}

\paragraph{What is ICL?} ICL is the ability of a pre-trained model to solve tasks without task specific fine-tuning \citep{radford2019language, brown2020language, olsson2022context}. Many large models have shown excellent ICL capabilities \citep{touvron2023llama,chowdhery2022palm}. This has shifted the research community's focus towards leveraging prompts to elicit zero-shot or few-shot responses from models. In a similar vein, the technique of chain-of-thought  (CoT) reasoning, as discussed in \citet{wei2022chain}, revealed that including a sequence of intermediate reasoning steps can enhance the reasoning skills of large language models. Yet, these abilities are emergent, i.e., it is primarily the larger models that exhibit them.  
However, recent studies question the belief that improvements in ICL result exclusively from increasing model sizes \citep{schaeffer2023emergent, du2024understanding}, suggesting that 
%emergent abilities 
%- abilities not seen in smaller-scale models but suddenly appear in larger-scale models -
%only appear as such due to the selection of evaluation metrics.
using discontinuous metrics like accuracy merely creates the illusion of emergent abilities, whereas employing continuous metrics %like loss 
shows gradual, predictable changes in model performance. 
%\newline

\paragraph{ICL in smaller language models.} It has been shown that the emergent abilities observed in larger models can be effectively transferred to smaller models through imitation learning or behavior cloning, where a larger language model such as GPT-4 serves as the ``teacher'', to generate synthetic datasets with instructions and explanations which can be used to train smaller language models, referred to as ``student'' models \citep{alpaca, peng2023instruction, mukherjee2023orca, magister-etal-2023-teaching}. This allows smaller models to leverage the capabilities of their larger counterparts. However, the primary drawback of such methods is that most of the knowledge acquired by the model is done in the pre-training stage and the student model copies the style of the teacher model but does not learn the reasoning capabilities employed by these large models \citep{gudibande2023false}.

An alternative strategy to enhance the capabilities of smaller models is through distillation from larger models, aiming to replicate the output probabilities and thus transfer the larger model's in-context learning or zero-shot abilities to their smaller counterparts \citep{timiryasov2023baby, gu2023minillm, xu2024survey}. This method forfeits one of the primary benefits of smaller language models, namely their reduced computational requirements, by necessitating the training of larger models.

% To improve the generation quality of small language models \citet{zhao2023babystories} pre-train language models on the babyLM data and fine-tune on a simpler dataset using reinforcement learning with human feedback and find that smaller language models (<120M) parameters do not benefit much from RLHF.

Prior work has also looked into pre-training small models with simplified data. For instance, \citet{huebner2021babyberta} pre-train an encoder language model with corpus that reflects the lexical exposure of children and find that smaller models can approximate the grammatical acquisition performance of larger models. \citet{deshpande2023honey} examined the effects of downscaling the modeled language during pre-training via vocabulary-based filtering, and showed that pre-training encoders as small as 1.25M parameters may demonstrate large benefits for downstream performance.

\citet{eldan2023tinystories} have demonstrated that coherency in text generation can be achieved by pre-training on a synthesized, simplified dataset generated from GPT-4. Notably, this dataset largely comprises of stories, presenting less diversity compared to the datasets typically employed for pre-training larger models. Similarly \citet{gunasekar2023textbooks} demonstrate improved performance in smaller models trained on a dataset combining filtered coding examples and synthetic textbook content for coding-related benchmarks. However, their approach, primarily focused on coding challenges, utilizes relatively large models and synthetic data. Their dataset filtering approach also relies on an auxiliary classifier for text exclusion.

%Recent interest in simplifying the pre-training data and training smaller models has resulted in challenges such as the BabyLM challenge \citep{warstadt2023findings} which focuses on a fixed data budget of 100M words for pre-training. It tries to simplify the pre-training data so as to mimic human acquisition of language in children and show that models trained on such a small datasets can also outperform models trained on larger datasets. 
%To this end, they create a pre-training dataset using child-directed speech, children’s literature, and transcribed dialogue containing a maximum of 100M words and show that models trained on such a small dataset can also outperform models trained on larger datasets. 
%However, their approach restricts the domain of pre-training data and exhibits less diversity compared  to standard pre-training corpora.

\section{Methodology}
\subsection{Language Simplification}
We create a simpler pre-training corpus by utilizing a vocabulary derived from the AO-Childes transcripts of child-directed speech \citep{huebner2021using}, as done by \citet{deshpande2023honey}. The core of this corpus is child-directed speech, which tends towards simpler linguistic structures. The vocabulary we use comprises 21,036 unique words, reflecting the lexical range typically found in language directed at children. Filtering existing pre-training corpora with this vocabulary thus results in a simpler pre-training dataset.

% from results of next section for better placement
% \input{tables/table_blimp_filtered_limited}

\subsection{Pre-training Data Collection}
To obtain high quality datasets with sufficient deduplication and diversity we leverage datasets used for pre-training large language models such as the  SlimPajama dataset \citep{cerebras2023slimpajama}. We begin by selecting samples from the train split of the SlimPajama pre-training corpus, then tokenize this text into distinct elements, such as words and symbols. We retain tokens that are either integers, special symbols, or belong to the AO-Childes vocabulary. 
%This process continues until we accumulate a minimum of 100 tokens in a chunk. 
This process continues until we accumulate a minimum number of tokens in a chunk. For the 22 Billion dataset the minimum number of tokens are set to 32 and for the 2.1 Billion dataset the minimum number of tokens are set to 100. 
We allow up to 1.5\% of these tokens to be out-of-vocabulary (OOV) words to maintain a simplified vocabulary and yet allow some linguistic variability. If the percentage of OOV words in a chunk exceeds this 1.5\% threshold, we conclude the current chunk and initiate a new one at the beginning of the next sentence. This approach ensures that each analyzed text chunk primarily consists of known vocabulary words, with a minimal presence of OOV words. Figure \ref{fig:method} illustrates our method. 

After filtering, our datasets consist of around 22 billion and 2.1 billion tokens, derived from various splits of the SlimPajama dataset. The distribution of the tokens on the 22 billion dataset across various splits of the SlimPajama dataset is detailed in Table \ref{tab:dataset_simple_22B}. For the 2.1 Billion dataset the distribution of tokens can be found in Table \ref{tab:dataset_simple} and \ref{tab:dataset_regular} in the appendix.

We computed the Zipfian Coeffcient by fitting a linear regression model on the logarithm of word frequencies and ranks of words and obtained a zipfian coefficient of -1.11 indicating %that the dataset follows Zipf's law quite closely. %
the dataset exhibits a distribution pattern typical of natural languages, where a small number of words are extremely common, while the majority are rare, thereby underscoring naturalistic quality of our dataset \footnote{This analysis was done on the 2.1B dataset}. We utilize this dataset for pre-training of models we label as ``simple'' models (henceforce, Simple). In contrast, for our ``regular'' models, we train them using data from the SlimPajama dataset, applying no filtering and maintaining a similar number of tokens.

% We refer to the models pre-trained with the simplified  pre-training data as ``simple models'' (henceforce, Simple), and the models pre-trained with the unconstrained standard data as ``regular models'' or ``pre-trained baselines''.

\section{Experimental Setup}

\subsection{Model Configurations}
We pre-train transformer-based models \citep{vaswani2017attention} by adapting the LLaMA architecture \citep{touvron2023llama} and vary key hyperparameters, such as the number of dimensions of the hidden representations (hidden size), number of hidden layers in the Transformer decoder (num layers) and the internal dimension of the MLP (intermediate size). We keep the base period of the RoPE embeddings \citep{su2023roformer} (rope\_theta) at 20.

We trained 2 models on 22 billion tokens: Simple 165M and Simple 100M.
%with results reported in subsections \ref{sec:zero_shot_5.1}, \ref{sec:zero_shot_5.2}, and \ref{sec:zero_shot_5.4}. 
We also trained 54 models on 2.1 Billion tokens. Of these, 36 models were used for zero-shot experiments in section \ref{sec:zero_shot} of the appendix, %\ref{sec:zero_shot_pa}, \ref{sec:zero_shot_blimp}, and \ref{sec:zero_shot_unfiltered}%
while the remainder were utilized for curve fitting analyses in section \ref{sec:power_laws}. The hidden sizes used in our experiments are [32, 64, 128, 256, 512, 1024], with layer counts of [2, 4, 8]. For the majority of zero-shot experiments, the intermediate size was set at four times the hidden size. However, for the experiments detailed in section \ref{sec:power_laws}, we used intermediate sizes that are twice the hidden size as well. These variations in hyperparameters produced models from 1 million to 165 million parameters, as detailed in Table \ref{tab:eval_loss_perplexity} in the appendix. For training, we utilized the Flash Attention mechanism introduced by \citet{dao2022flashattention}.

\subsection{Model Pretraining}
We train a tokenizer using Byte Pair Encoding (BPE) \citep{sennrich2015neural} on the filtered dataset. We use a vocabulary size of 15000. All simple models are pre-trained on a causal language modelling objective for a single epoch on the simplified data derived from various splits of SlimPajama dataset. We train two sets of models one set on 22 Billion tokens and another set on 2.1 Billion tokens. The models trained on 22 Billion tokens have an effective batch size of 512 and context lengths of 1024 with model parameters being updated 41697 times. We use a cosine learning rate scheduler with warmup and use peak learning rates in the range of $6 \times 10^{-4}$ to $1 \times 10^{-3}$ (learning rates are chosen based on model size) and decay the learning rate down to 13\% of the peak learning rate and a perform warmup for 4000 steps. The models trained on 2.1 Billion tokens have an effective batch size of 4096 and context lengths of 128. The model parameters are updated a total of 3972 times with cosine learning rate scheduler with warmup and a peak learning rate of $2.8 \times 10^{-3}$, we decay learning rate upto 11\% of the peak learning rate and perform warmup for 1000 steps.

% We train models using 2.1 Billion tokens and 
% 21.5 Billion tokens, with an effective batch size of 4096 and context lengths of 128 and  
% 1024 tokens (524288 tokens per batch), we update model parameters a total of %3972%
% 41697 times. We use a cosine learning rate scheduler with warmup and use a peak learning rates in the range of $6 \times 10^{-4}$ to $1 \times 10^{-3}$ %$2.8 \times 10^{-3}$%
% and decay the learning rate down to %11\%%
% 13\%of the peak learning rate and a warmup for 4000 steps. 
We utilize the AdamW optimizer for updating the parameters, with the $\beta_1$, $\beta_2$, and $weight\_decay$ values of 0.9, 0.95, and 0.1 respectively. We use gradient clipping of 1.0. We conduct pre-training of all models on 2 RTX3090. For training regular models, we use a dataset consisting of 2.1 Billion tokens, and use the same hyperparameters as those used for training simple models with the same token count.

\subsection{Model Evaluation and Datasets}
We evaluate all pre-trained models for their zero-shot and few-shot ICL abilities using the language model evaluation harness from EleutherAI \citep{eval-harness}, which is a framework designed to perform zero-shot and few-shot evaluations. The datasets we use include: Benchmark of Linguistic Minimal Pairs for English (BLiMP) \citep{warstadt2020blimp} to assess the models' capability in understanding linguistic features. BLiMP is comprised of 67 individual datasets, each containing a pair of sentences: one grammatically correct, and the other incorrect. These pairs are designed to assess the models' proficiency in recognizing morphological, syntactic, and semantic aspects of language. Additionally, we use the Physical Interaction Question Answering (PIQA) \citep{Bisk2020} to measure the performance of lanuguage models on questions requiring physical common sense, AI2 Reasoning Challenge (ARC-Easy) \citep{allenai:arc} which measures reasoning abilities of Language Models, Choice of Plausible Alternatives (COPA) \citep{roemmele2011choice} which evaluates common sense causal reasoning of language models, Microsoft Research Paraphrase Corpus (MRPC) \citep{dolan2005automatically} which evaluates if the language model can identify if a pair of sentences constitutes a paraphrase, RTE which evaluates if language models can identify entailment and non-entailment, MultiGenre Natural Language Inference (MNLI) \citep{williams2017broad} which provides a more diverse and challenging dataset for natural language inference, and Stanford Sentiment Treebank (SST-2) \citep{socher2013recursive} which evaluates the  model's fine grained understanding of sentiment. 

In reporting our findings, we differentiate between the entire downstream dataset called the ``unfiltered dataset'' or ``standard dataset'' and the filtered subset termed the ``filtered'' or ``simplified dataset''. The simplified version of each downstream task consists of instances using only the words from the AO-Childes lexicon (with the addition of digits and special symbols), thereby mitigating potential distributional shifts between pre-training and testing. 

The primary aim of our study was to understand if smaller models could achieve performance improvements similar to those observed in larger models, but when modeling a simplified language. We would like to emphasize that this choice is motivated by our research goal: to understand whether the lack of emergent abilities in smaller pre-trained models is merely a question of model capacity, and whether reducing the problem (down-scaling the language) would allow us to see similar abilities emerge in smaller models. This logic is what directly motivates our downstream evaluation with standard datasets filtered down to imitate the constrained language setting. However, we recognize the value of evaluating model performance on unfiltered datasets for comparison. To this end, we have included the performance of simple models on unfiltered datasets as well.

% Assessment across the entirety of the dataset serves as a measure for evaluating the simple models' generalization ability.
% We refer to the models pre-trained with the simplified  pre-training data as ``simple models'' (henceforce, Simple), and the models pre-trained with the unconstrained standard data as ``regular models'' or ``pre-trained baselines''.

% We pre-train our models with context length of 128 to increase the number of contiguous spans that can be extracted from the regular dataset. However, datasets such as PIQA and ARC Easy contain examples that span more than the pre-trained context length. To extend context window sizes beyond 128, we use Position Interpolation \citep{chen2023extending} on PIQA and ARC Easy datasets. We use a scaling factor of 8 which allows to have context window of 1024. For more information regarding Position Interpolation please refer to Appendix \ref{sec:appendix_rope}.

% zero shot

\begin{table*}
\centering
\footnotesize
\begin{tabular}{lcccccccc|c}
\toprule
\textbf{Model} & \textbf{COPA} & \textbf{MRPC} & \textbf{RTE} & \textbf{MNLI} & \textbf{ARC-Easy} &  \textbf{BLiMP} & \textbf{PIQA} &  \textbf{SST-2} &  \textbf{Average}\\
\midrule
{Pythia 1B} & 0.72 & 0.68 & 0.53 & 0.34 & 0.57 & 0.84 & 0.71 & 0.50  & 0.61 \\
{Pythia 1B}\textsuperscript{\textdagger} & 0.68 & 0.67 & 0.50 & 0.35 & 0.56 & 0.83 & 0.70 & 0.65  & 0.62  \\
\midrule
{Pythia 410M} & 0.70 & 0.52 & 0.53 & 0.34 & 0.52 & 0.84 & 0.67 & 0.51 & 0.58\\
{Pythia 410M}\textsuperscript{\textdagger} & 0.71 & 0.38 & 0.50 & 0.34 & 0.52 & 0.83 & 0.66 & 0.66 & 0.58\\
\midrule
{Pythia 160M} & 0.63  & 0.67 & 0.52 & 0.35 & 0.44 & 0.80 & 0.62 & 0.51 & 0.57 \\
{Pythia 160M}\textsuperscript{\textdagger} & 0.61 & 0.67 & 0.50 & 0.37 & 0.41 & 0.80 & 0.61 & 0.36 & 0.54 \\
\midrule
\midrule
{OPT 1.3B} & 0.79 & 0.66 & 0.51 & 0.36 & 0.57 & 0.86 & 0.72 & 0.82  & 0.66  \\
{OPT 1.3B}\textsuperscript{\textdagger} & 0.73 & 0.62 & 0.57 & 0.37 & 0.59 & 0.85 & 0.70 & 0.90  & 0.67  \\
\midrule
{OPT 350M} & 0.72 & 0.68 & 0.52 & 0.34 & 0.44  & 0.85 & 0.65 & 0.62 & 0.60 \\
{OPT 350M}\textsuperscript{\textdagger} & 0.73 & 0.67 & 0.71 & 0.36 & 0.45 & 0.84 & 0.63 & 0.71 & 0.64\\
\midrule
{OPT 125M} & 0.66 & 0.68 & 0.50 & 0.34 & 0.44 & 0.83 & 0.63 & 0.53 & 0.58 \\
{OPT 125M}\textsuperscript{\textdagger} & 0.61 & 0.67 & 0.50 & 0.34 & 0.47 & 0.83 & 0.63 & 0.42 & 0.56 \\
\midrule
\midrule
{Simple 165M} & 0.73 & 0.68 & 0.56 & 0.33 & 0.35 & 0.71 & 0.63 & 0.49 & 0.56 \\ 
{Simple 165M}\textsuperscript{\textdagger} & 0.83 & 0.67 & 0.79 & 0.35 & 0.42 & 0.76 & 0.65 & 0.64 & 0.64 \\ 
\midrule
{Simple 100M} & 0.66  & 0.68 & 0.52 & 0.33 & 0.34 & 0.72 & 0.62 & 0.60 & 0.56 \\
{Simple 100M}\textsuperscript{\textdagger} & 0.68 & 0.58 & 0.64 & 0.35 & 0.43 & 0.78 & 0.64 & 0.58 & 0.59 \\
\midrule
\midrule
{Random Chance} & 0.50 & 0.50 & 0.50 & 0.33 & 0.25 & 0.50 & 0.50 & 0.50 & 0.45\\
\bottomrule
\end{tabular}
\caption{Zero-shot accuracy of pre-trained Pythia and OPT models vs. models trained on simplified language. Models are evaluated on both standard and vocabulary-filtered datasets. Results on vocabulary-filtered datasets are marked with  \textsuperscript{\textdagger}. Our findings indicate that the simplified models demonstrate superior zero-shot performance on vocabulary-filtered datasets, achieving higher average scores across these datasets compared to the average scores of significantly larger Pythia pre-trained models.}
\label{tab:zero-shot-filtered-unfiltered}
\end{table*}
\section{Results} 

Our goal is to understand if the absence of emergent abilities in smaller pre-trained models is simply a matter of model capacity and whether simplifying the problem, i.e., downscaling the language, would allow these abilities to emerge in smaller models. To this end, we evaluate models trained on simplified data against both filtered and standard evaluation datasets. We compare our models with Pythia \citep{biderman2023pythia} and OPT \citep{zhang2023opt} (models up to 1.3B parameters) to determine if downscaling the language facilitates emergent capabilities to  occur much earlier.

% We compare models trained on simplified data against those trained on regular data to substantiate our hypothesis: small models trained on simplified data outperform their counterparts trained on a standard data on grammar and simple tasks.

Table \ref{tab:eval_loss_perplexity} in the appendix shows the perplexity for different-sized models trained on both simple and regular datasets. The simple dataset is derived from a subset of the SlimPajama dataset, where the text has been filtered to limit vocabulary complexity. In contrast, the regular dataset uses the original, unaltered text from the same source. A separate test set, similar in distribution to the training data, was used to evaluate perplexity in each set of models. The reported results reflect the performance of each model trained in an identical training regimen with the same number of training steps. We find that as the model size increases, its ability to accurately predict and understand the held-out test set also improves, as evidenced by decreasing perplexity on both the simple and regular models. Furthermore, the perplexity metrics indicate that at this scale, simple models are able to learn the simple language much better. Simple models in the range of 1-165M parameters achieve perplexity of 92.00 - 20.59 on the simple dataset. In contrast, when regular models are trained on a regular dataset they achieve perplexity in the range of 193.20 - 28.97 on the regular dataset.\footnote{These results are reported on models trained on 2.1B dataset}

% Moving it for better placement
\begin{figure*}
    \centering
    \includegraphics[width=\textwidth]{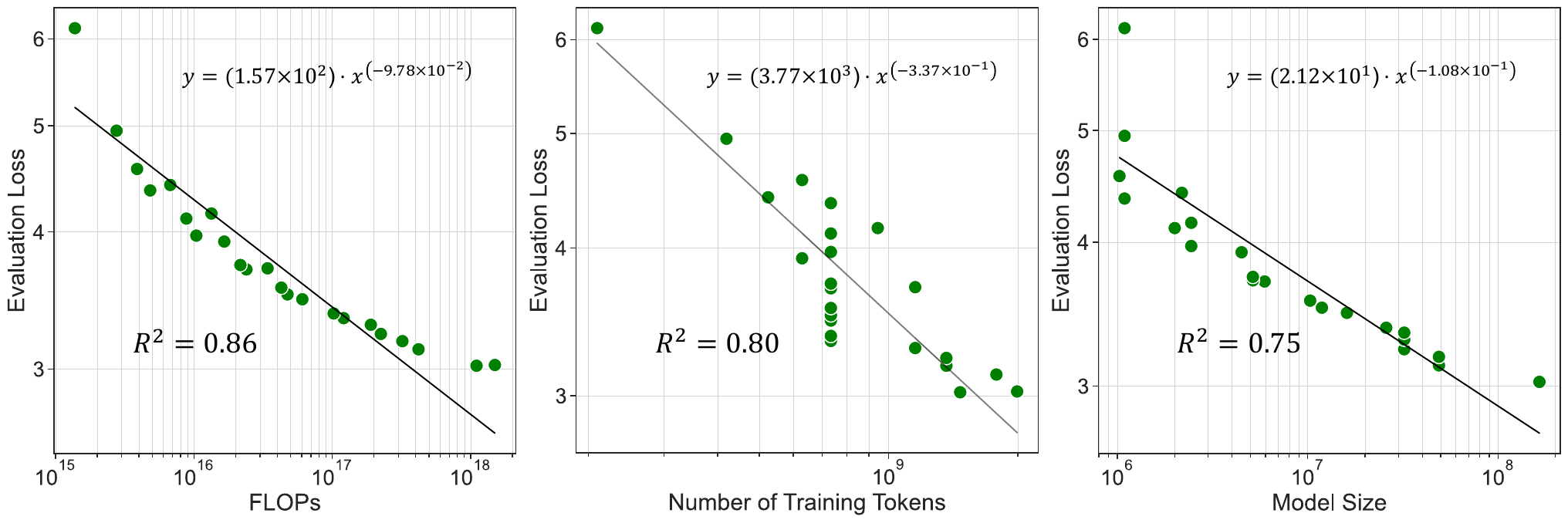}
    \caption{Here we present our curve-fitting results. The green dots represent the compute-optimal instances found in our experiments and the black solid line represents the fitted power curve of the form $y=A \cdot x^{B}$. In each subfigure, we provide the optimal values of $A$ and $B$, and the goodness of fit ($R^2$). Starting from the left, we present the relationship between the evaluation loss (y-axis) and FLOPs (left subfigure), pre-training data (center subfigure) size, and model size (right subfigure), respectively. All $R^2$ values are over 0.74 and we observe the best fit for the left subfigure (loss vs. FLOPs). In the test range of values for model size and data size, we observe that loss value reduces faster per unit change in model size.}
    \label{fig:scaling_laws_panel}
\end{figure*}
\subsection{Do simple models perform better in \textit{zero-shot} settings?}
\label{sec:zero_shot_5.1}
In-context learning, as defined by \citet{brown2020language}, enables models to apply knowledge gained during pre-training to new tasks without requiring fine-tuning on task-specific datasets. We evaluate the ICL capabilities of our models, focusing on their zero-shot performance across a range of tasks, including COPA, MRPC,  RTE, MNLI, SST-2, PIQA, ARC-Easy, and BLiMP. These tasks are analyzed using both standard and vocabulary-filtered datasets.

Table \ref{tab:zero-shot-filtered-unfiltered} presents zero-shot performance for different models, including pre-trained Pythia models (1B, 410M, 160M), OPT models (1.3B, 350M, 125M), and models trained on simplified language (165M and 100M), which we will refer to as Simple models. 
The Simple models perform better on vocabulary-filtered downstream tasks than on the corresponding unrestricted standard versions of the same tasks. This is to be expected, since these models are not exposed to unrestricted language during training.

Since the larger models used in this study are pre-trained on unrestricted language, we expect them to be able to handle tasks using simplified language, as the latter is a subset of their training data. 
%both standard and simplified datasets effectively.
Interestingly, we see that Simple 165 model outperforms Pythia 1B model on simplified downstream data (0.64 vs. 0.62 average performance), suggesting that modeling a restricted language allows smaller models to achieve stronger-than-expected zero-shot capabilities.

A curious comparison arises between the performance of small models on simplified tasks and the performance of larger models pre-trained on standard language on the corresponding standard versions of the same tasks. In this situation, there is no distribution shift between training and testing. If both the model size and language complexity are downscaled appropriately, we expect to see similar performance figures. 
However, we see that the Simple 165M model performs better on simplified downstream data than the Pythia 1B on standard datasets (0.64 vs. 0.61 average performance), despite being approximately six times smaller and seeing substantially less data. We see a similar trend with OPT model family, where the Simple 165M model does better on the simplified downstream data than OPT 350M model on standard datasets (0.64 vs. 0.60 average performance). 
%This comparison underscores that simplifying vocabulary, allows smaller models to enhance their zero-shot ICL capabilities.

We also report the performance of small models trained on a much smaller amount of data (2.1B tokens), comparing regular and restricted pre-training. For detailed performance comparison on the BLiMP benchmark, PIQA, and ARC-Easy datasets across different model sizes, please refer to the appendix \ref{sec:zero_shot}. 
As expected, pre-training smaller language models on simpler data leads to better downstream task performance.
Consistent with \citet{deshpande2023honey}, we see above random performance of models as small as 1M parameters.
Figure \ref{fig:combined_acc} in the appendix, shows the zero-shot task accuracy with respect to the hidden size and number of layers.

%and their regular counterparts on the BLiMP benchmark, PIQA, and ARC-Easy datasets. For detailed performance comparison across different model sizes, please refer to the appendix \ref{sec:zero_shot}. 
%Each model size and layer combination has two rows of data: the first for simple models, and the second for regular models. 
%A notable trend is the superior average performance of simple models compared to their regular counterparts, confirming that pre-training smaller language models on simpler data leads to better downstream task performance. 
%Additionally, an increase in model capacity generally leads to improved performance. 
%Also, we see non-random performance even in our smallest models containing around 1 million parameters. We notice a similar trend on the average BLiMP score and PIQA scores with the unfiltered datasets as well. 
%Results, categorized by model size (hidden layers and number of layers), are provided in appendix \ref{sec:zero_shot}. Figure \ref{fig:combined_acc} in the appendix, shows the zero-shot task accuracy with respect to the hidden size and number of layers. These results are on models trained with 2.1B tokens.

%  However, the ICL capabilities of smaller models are predominantly restricted to simpler tasks. 
% In more complex tasks, these models exhibit performance close to random indicating that you do need greater capacity to get better performance in a zero-shot setting. 

\subsection{Do Simple models perform better in \textit{few-shot} settings?}
\label{sec:zero_shot_5.2}

%We also compare few-shot, rather than zero-shot, performance of simple and standard pre-trained models.
In addition to zero-shot performance, we also compare the few-shot performance of simple and standard pre-trained models.
We evaluate the performance of the Simple 165M and Simple 100M models, which are pre-trained on a simplified vocabulary, against the Pythia baselines (160M, 410M, and 1B).
This evaluation uses few-shot prompting using examples from vocabulary-filtered downstream data.
We report the models' average performance across the following datasets: COPA, MRPC, RTE, MNLI, ARC-Easy, PIQA, and SST-2. The results for each dataset are averaged over three runs, with each run using different task examples in the context.

From the results in Table \ref{tab:few-shot}, we observe no significant improvement in performance with an increased number of in-context examples.
This is in line with previous findings for language models of similar sizes \citep{brown2020language}. This suggests that the smaller model sizes of 100M or 165M may not be adequate to fully demonstrate the few-shot ICL capability in the downscaled language setting. 
We also believe that the simplified-data models we investigated likely lacked the scale necessary to exhibit emergent abilities such as chain-of-thought prompting \cite{wei2022chain}. Just as models smaller than 10B parameters trained on unrestricted language actually perform worse with CoT prompting \cite{wei2022chain}, our simplified-data models may also require greater scale to exhibit such capabilities.

% Similar to \citet{mitra2023orca}, we observe that few-shot prompting does not significantly aid these smaller models. This could be due to the limited context window inherent in the training of these models. Although we employ position interpolation to extend the context length during testing, this might not sufficiently compensate for the initial training limitations or these models require greater capacity to effectively learn from few-shot examples. Detailed few-shot results can be found in Appendix \ref{sec:appendix_few_shot}

% few shot

\begin{table}
\centering
\tiny
\begin{tabular}{lccccc}
\toprule
\textbf{Model} &\textbf{0-shot}  &\textbf{1-shot} & \textbf{2-shot} & \textbf{3-shot} & \textbf{4-shot} \\
\midrule
{Pythia 1B} & 0.59 & 0.57 & 0.57 & 0.58 & 0.60  \\
{Pythia 410M} & 0.54 & 0.55 & 0.53 & 0.56 & 0.54 \\
{Pythia 160M} & 0.50 & 0.50 & 0.48 & 0.51 & 0.50  \\
\midrule
{Simple 165M} & 0.62 & 0.56 & 0.56 & 0.54 & 0.54  \\ 
{Simple 100M} & 0.56 & 0.56 & 0.56 & 0.56 & 0.55  \\
\bottomrule

\end{tabular}
\caption{Average few-shot results across different vocab-filtered tasks such as COPA, MRPC, RTE, MNLI, ARC-EASY, PIQA, SST. Our results reveal no discernible trend in the few-shot learning results, suggesting that larger models are required to observe the emergence of few-shot in-context learning capabilities.}
\label{tab:few-shot}
\end{table}
% \subsection{Does vocabulary based filtering help improve performance on grammar related tasks?}
% \label{sec:zero_shot_blimp}

% Table \ref{tab:blimp_comp_contrastive_limited} 
% TODO refer to the complete table

% However, in some sub-categories such as Binding and Island Effects, regular models outperform simple models.

\subsection{Do simple models obey power laws?}
\label{sec:power_laws}
We fit a power curve of the form $L = A \cdot x^{B}$, to predict the cross-entropy loss (L) based on the compute cost (C), data size (D), and model size (M), separately. For curve-fitting, we consider only the 25 compute-optimal instances found for 25 bins of the FLOPs values and utilize $R^2$ value to assess the goodness of fit. We adopt the formula presented by \citet{deshpande2023honey} to calculate the FLOPs values which considers the embedding parameters while calculating FLOPs unlike \cite{kaplan2020scaling, hoffmann2022training}. Similar to \citet{kaplan2020scaling, hoffmann2022training} we observe that the upstream performance (pre-training cross-entropy loss on validation split of the data) is fairly predictable with $R^2$ value of 0.86, 0.80, and 0.75, for compute cost, data size, and model size, respectively. We also observe that improvement in the loss value is faster for the model size compared to the data size. \footnote{The results are reported on models trained on 2.1B tokens}
%with the exponential coefficients being $-0.07$, $-0.06$, and $-0.08$ with the goodness of fit as measured by r2 score being 0.91, 0.96, and 0.76. This shows that in the low parameter and constrained data regime our models follow the scaling laws shown by bigger models.

% TODO do we want this now?
% \input{tables/table_generalization}

%TODO is it a good idea to have the filtered and unfiltered table as separate?
% \input{NAACL_2024_camera_ready/tables/table_zero_shot_unfiltered.tex}

\subsection{Do Simple Models yield good generations?}
\label{sec:zero_shot_5.4}
\begin{table*}
\centering
% \begin{tabular}{p{0.15\textwidth} p{0.55\textwidth} p{0.15\textwidth}}
% \begin{tabular}{lllll}
\footnotesize
\begin{tabular}{p{0.17\textwidth} p{0.17\textwidth} p{0.17\textwidth} p{0.17\textwidth} p{0.17\textwidth}}
\toprule
\textbf{Prompt} & \textbf{Simple-165M} & \textbf{Pythia-160M} & \textbf{Pythia-410M} &  \textbf{Pythia-1B} \\
\midrule
I saw my friend Joe sitting in lobby today. I kept him company, as he is a lonely old man. He told me he had just listened to Beethoven's Ninth. I talked to him for an hour. I left him in the lobby and told him I would see him soon. & I went to check on him and he was just sitting there waiting. Then I got to see a few people who were with him that night. He had been going into his room after his accident. I went down to the lobby to see what was & He seemed to think I was the right person. We talked on the telephone for two hours. Then I came back to the room where we had been sitting at the dinner table and went out. There was a young woman sitting at the opposite end of & We went to his room. He was asleep, lying in bed, as I was. He did not wake up and I did not want him to. My friend Joe told me this afternoon that he had been on the way to visit his & He seemed to be enjoying himself. I did not leave him alone as he has been known to do. After his hour with me he decided to have a drink. He seemed very pleased with himself. I left him alone. I hope he \\
% \midrule
% Dan wanted a pet for Christmas.He told his dad.His dad listened, but didn't say anything.So on Christmas morning, And got a wonderful surprise.He received a puppy with a shiny bow on it's head! & Now it's a new puppy. So excited, and so grateful. I hope we could all be together again. Because it's Christmas! You're all I could ever ask for! But that's because you're all I & It's been a real joy watching it grow, so it's a big happy surprise!The puppy looks a bit scary to me, but I'm really impressed.His mother is very supportive of him, but he's not happy with the puppy either & It was soooo cute.His Dad said to And...to...to say...hello!And so he did... & 
% The puppy grew to love all the people who had invited him. It would get on his head and run after them. And chase them down! It would have gotten them in a lot of trouble!

% One year, when the neighbors \\
\midrule
Todd was hungry. He did not have anything to cook at his house. He decided he need to go buy something to cook. On the way to the store Todd decided to make hamburgers. Todd buys everything he needed and goes home and cooks. & When he arrived, he saw his friends there cooking ham burgers for them. He was already thinking about dinner. He had decided he would make a ham burger. T odd looked at his friends and he knew that he had to do something about this. & When he's done cooking, he takes the ham off the rack. He makes the ham into a turkey, then adds a little salt and pepper. Todd then steaks them with chicken.The ham goes through the oven in a double skillet and & He had some leftovers left. He wanted to do this for the rest of his life.He made up for his lunch with some chips. His mother came home from work. She told him to make her hamburger. She had to eat & When he got back home he took all the food he needed. He eats all the food and it was too late to cook something else now he was back at the store. Todd decides to go in the store again. There he finds out that \\
\bottomrule
\end{tabular}
\caption{Comparison of the prompt completions generated by the Simple 165M model trained on vocabulary-filtered simplified pre-training dataset and Pythia pre-trained baselines (160M, 410M, 1B). For decoding of all models we set temperature to 1.0 and employ nucleus sampling with top\_p set to 0.9. The maximum number of new tokens are set to 50.}
\label{tab:generations}
\end{table*}

We analyze text continuations, on prompts sampled from TinyStories \cite{eldan2023tinystories} and ROCStories \cite{mostafazadeh2016corpus}, using the 165M simple model and Pythia baselines (160M, 410M, 1B). We sample 25 different prompts from both these datasets randomly. We choose prompts from these datasets so as to keep the prompts simple enough for the model trained on vocabulary-filtered pre-training dataset to understand. For decoding of all models we set temperature to 1.0 and employ nucleus sampling \citep{holtzman2019curious} with top\_{p} set to 0.9. The maximum number of new tokens are set to 50. 

\begin{figure}
    \centering
    \includegraphics[width=0.5\textwidth]{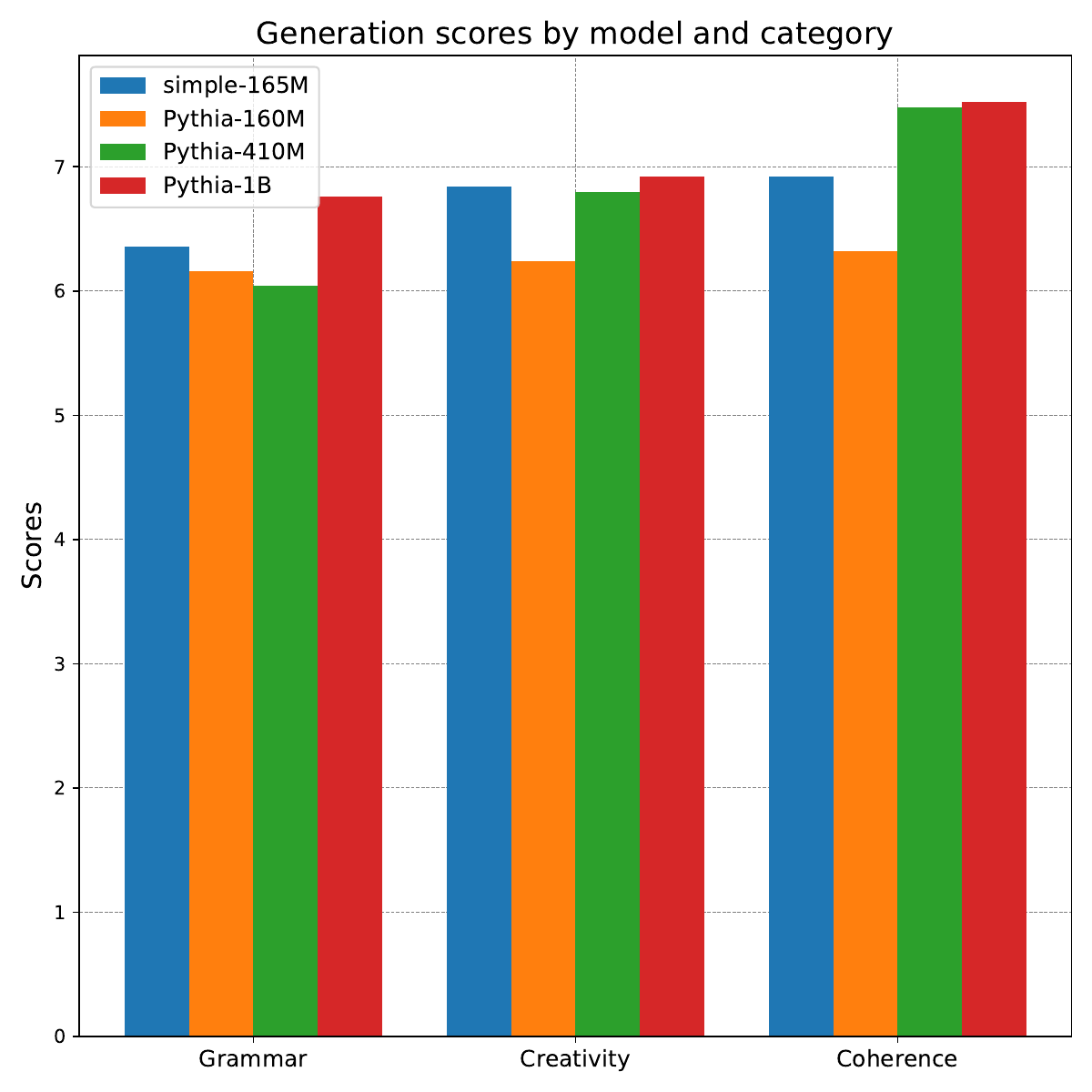}
    \caption{Comparative Analysis of Text Quality from Various Generative Models as evaluated by GPT-4. Models pre-trained on a vocabulary-simplified dataset produce outputs that are grammatical, creative, and coherent, and are comparable to those from larger models trained on the complete language dataset.}
    \label{fig:scores_generations}
\end{figure}

Table \ref{tab:generations} shows few initial prompts and generations from simple models and different baselines. Similar to \cite{eldan2023tinystories} we evaluate the generations with GPT-4, to assign scores ranging from 1 to 10 with 10 being the highest  for different aspects of the generated text such as grammar, creativity, and coherence. We plot the average scores across all completions for each model as depicted in  Figure \ref{fig:scores_generations}. From the figure it can be seen that the simple model performs comparably to the Pythia 410M model in terms of grammar and creativity and the simple model outperforms Pythia 160M model in terms of coherence.

% \subsection{Do simple models generalize well?}
% \label{sec:zero_shot_unfiltered}
% To determine the generalization capabilities of simple models, we conducted a zero-shot performance evaluation on the unfiltered PIQA, ARC Easy and BLiMP datsets. Given that these models were initially trained on a vocabulary-filtered dataset, the unfiltered dataset includes out-of-distribution samples. As seen in Table \ref{tab:generalization}, we notice a decline in performance relative to the vocabulary-filtered dataset; however, the observed peroformance is still above random levels even in our smallest sized models. Hence we can conclude that simple models do generalize well on out-of-distribution data.

\section{Conclusion}

In our study, we explored the impact of simplifying pre-training data on the performance of small generative models, specifically those with fewer than 165 million parameters. Our primary focus was to assess whether these models exhibit emergent abilities, notably zero-shot learning — the capability to have non-random performance on tasks without explicit prior training. To this end, we evaluated a series of models, each varying in hidden size and the number of layers, and measured their zero-shot performance across different tasks. Our findings reveal that smaller models operating in a %down-scaled% 
simplified language regime indeed demonstrate enhanced zero-shot learning abilities on vocabulary-filtered datasets, and outperform larger baseline models trained on full language on standard datasets. This suggests that the complexity of training data is a crucial factor in the development of zero-shot learning capabilities in smaller models.

We expect future work to investigate the model and data scales at which other emergent abilities (such as few-shot ICL and CoT reasoning) appear when modeling a reduced-scale language.
Exploring the potential of instruction fine-tuning in   models trained with simplified language is another interesting direction to pursue in future work.
%It should also explore the potential of instruction-fine-tuning in models trained with simplified language data. 
%We aim to determine if this approach enhances the instruction-following capabilities which could result in more efficient model generations, particularly in low-resource computing environments.

\section{Limitations}
We adopt vocabulary reduction for simplifying the language we model and do not explore other possible ways of simplification such as sentence structure simplification, data pruning, or curriculum learning. The adopted vocabulary-based data filtration also leads to different distributions of sequence length and word frequencies compared to regular English data. Hence, our findings should be considered within the distributional properties of our pre-training data. We extend the pre-trained model capabilities to process longer sequences by utilizing position interpolation method \cite{chen2023extending}.
%Keeping the upper bound of 165 million trainable parameters, we explored various models by changing the architectural hyperparameters such as hidden size, intermediate size, and number of hidden layers. 
We train all models with only the causal language modeling task and do not consider distillation or model pruning as a means of developing smaller models. For evaluating the pre-trained models, we focus only on the in-context learning abilities. Hence, we keep the finetuning experiments out of the scope of our study. We further note that instruction tuning of models may considerably affect the ICL abilities. However, we keep the investigation of the effect of instruction tuning on ICL abilities for future work.

\section{Acknowledgement}
We would like to thank Namrata Shivagunde for her insightful suggestions on various aspects of experiments.
%This work was funded in part by 
%We investigate various model sizes by adjusting the hidden layer size and the number of layers, ensuring that the models have fewer than 165 million parameters to minimize computational costs. We train all our models on LLaMA architecture. The model's ability to follow instructions could be enhanced through instruction fine-tuning. Additionally, we filter the vocabulary to mainly include terms from a child-centric lexicon leading to fewer contiguous token spans. Therefore, we choose a context length of 128 in during pre-training. Post pre-training, we can offset these limitations to a certain extent by using position interpolation.

\bibliography{anthology,custom}
\bibliographystyle{acl_natbib}

\appendix

\section{Data distribution for the Simple and Regular Datasets with 2B tokens}

The data distribution corresponding to the simple and regular datasets containing 2B tokens can be found in \ref{tab:dataset_simple} and \ref{tab:dataset_regular} respectively.

\section{Modeling simple and regular language}
Table \ref{tab:eval_loss_perplexity} shows the perplexity metrics across different simple and regular models trained on 2.1B tokens. Perplexity of simple  models are evaluated on the  held out test set of simple pre-training data whereas perplexity of regular models are  evaluated on the held out test set of both simple and regular pre-training data. It can be seen that smaller models are better able to model simple data compared to regular data.

\section{Additional Zero-shot Evaluation Results}
\label{sec:zero_shot}
% from results for better placement
\begin{figure*}[h!]
    \centering
    \begin{subfigure}{0.45\textwidth}
        \includegraphics[width=\linewidth]{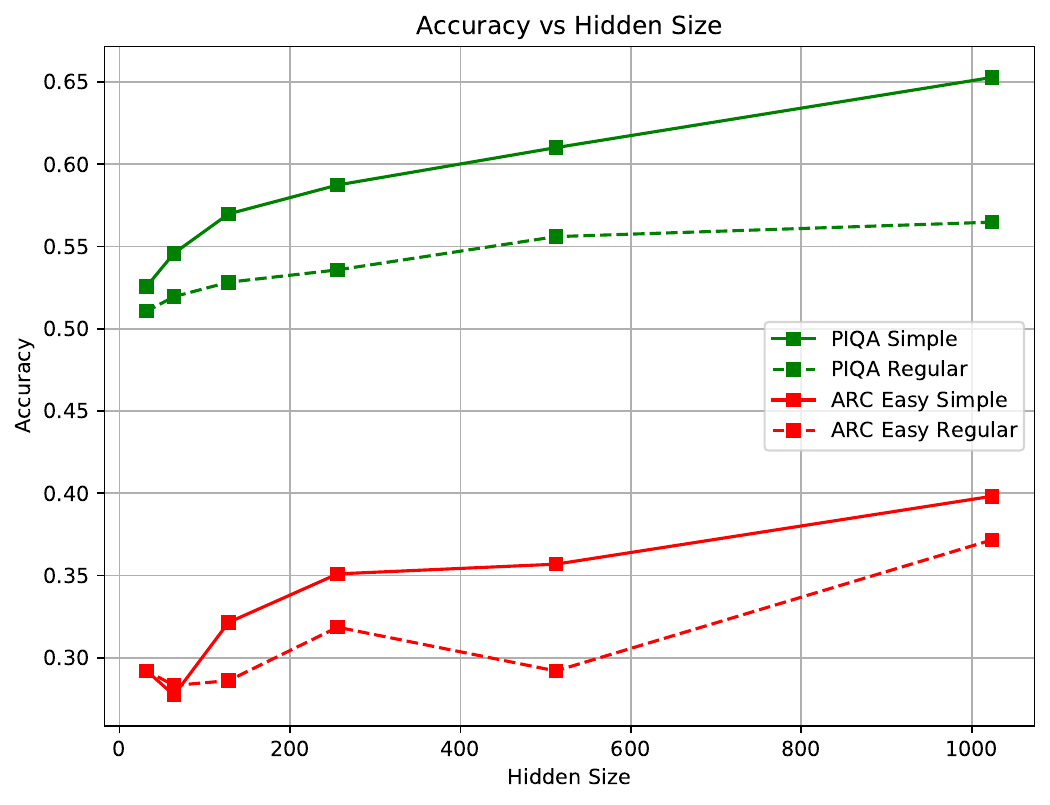}
        \caption{Variation in accuracy for hidden sizes ranging from 32 to 1024 on both simple and regular models. For each hidden size, we choose the model with 8 layers.}
        \label{fig:acc_vs_hidden}
    \end{subfigure}
    \hfill
    \begin{subfigure}{0.45\textwidth}
        \includegraphics[width=\linewidth]{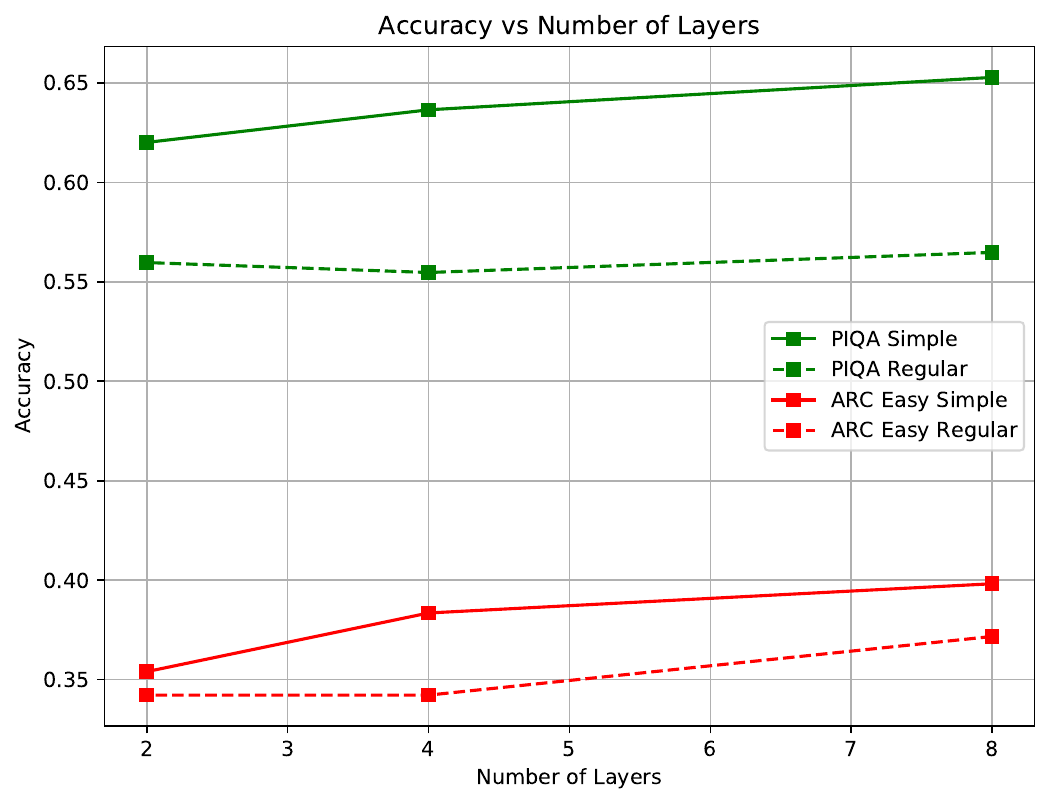}
        \caption{Variation in accuracy for layers ranging from 2 to 8 on both simple and regular models. For each layer we choose the model with a hidden size of 1024.}
        \label{fig:acc_vs_layer}
    \end{subfigure}
    \caption{Model performance different hidden layer sizes and number of layers on PIQA and ARC-Easy datasets. Both simple and regular models are compared on the filtered evaluation datasets. We observe that the simple models consistently outperform their regular counterparts across different model configurations.}
    \label{fig:combined_acc}
\end{figure*}

Figure \ref{fig:combined_acc} shows that the task accuracy improves with an increase in the capacity of the model for both regular and simple models. Simple models generally outperform their regular counterparts on simpler tasks such as  PIQA and ARC-Easy when evaluated on the filtered dataset.

% \input{tables/table_dataset_stats}
% \label{sec:appendix}
% \input{tables/table_few_shot_filtered}

% dataset dist

\begin{table}[t!]
\centering
\footnotesize
\begin{tabular}{lrr}
\toprule
\textbf{Split} & \textbf{Percentage of} & \textbf{Number of}\\
\textbf{} & \textbf{tokens} & \textbf{tokens (mil)}\\
\midrule
{C4} & {24.92\%} & {530.877}  \\
{GitHub} & {0.42\%} & {885.322}\\
{Commoncrawl} & {18.42\%} & {390.694} \\ 
{StackExchange} & {3.05\%} & {649.524} \\ 
{Wikpedia} & {0.09\%} & {186.751}\\
{ArXiv} & {2\%}  & {431.555}\\
{Books} & {51.2\%} & {1090.048}\\ 
\midrule
{total} & {100\%} & {2130.447} \\
\bottomrule

\end{tabular}
\caption{Data source distribution for the simplified pre-training dataset derived from SlimPajama.}
\label{tab:dataset_simple}
\end{table}
% dataset dist regular model pretraining

\begin{table}[t!]
\centering
\footnotesize
\begin{tabular}{lrr}
\toprule
\textbf{Split} & \textbf{Percentage of} & \textbf{Number of}\\
\textbf{} & \textbf{tokens} & \textbf{tokens (mil)}\\
\midrule
{C4} & {26.27\%} & {560.18}  \\
{GitHub} & {4.86\%} & {103.60}\\
{Commoncrawl} & {52.73\%} & {1124.51} \\ 
{StackExchange} & {3.15\%} & {67.16} \\ 
{Wikpedia} & {4.35\%} & {92.81}\\
{ArXiv} & {4.62\%}  & {98.55}\\
{Books} & {4.02\%} & {85.84}\\ 
\midrule
{total} & {100\%} & {2132.64} \\
\bottomrule

\end{tabular}
\caption{Data source distribution for the standard SlimPajama pre-training dataset used to train regular models.}
\label{tab:dataset_regular}
\end{table}
% training curves

\begin{table*}[ht]
% \begin{footnotesize}
\centering
\footnotesize
\begin{tabular*}{\textwidth}{@{\extracolsep{\fill}}cccccccc}
\toprule
% \textbf{Model} & \textbf{Anaphor} \\ \textbf{Agr.} & \textbf{Agr. Structure} & \textbf{Binding} & \textbf{Control/Raising} & \textbf{D-N Agr.} & \textbf{Ellipsis} & \textbf{Filler-Gap} & & \textbf{Irregular Forms}\\
% \makecell{{LLama} \\ {Unfiltered}} 
\makecell{\textbf{Model} \\ {\textbf{Size (M)}}} & \makecell{\textbf{Hidden} \\ {\textbf{Size}}} & \makecell{\textbf{Num } \\ {\textbf{Layers}}} & \makecell{\textbf{Int.} \\ {\textbf{Size}}} & \makecell{\textbf{PPL Simple} \\ {\textbf{(Simple Dataset)}}}  & \makecell{\textbf{PPL Regular} \\ {\textbf{(Regular Dataset)}}} &
\makecell{\textbf{PPL Regular} \\ {\textbf{(Simple Dataset)}}}  \\
\midrule
164.96 & 1024 & 8 & 4096 & 20.59 & 28.97 & 33.85 \\
97.84 & 1024 & 4 & 4096 & 24.59 & 31.13 & 35.49\\
64.28 & 1024 & 2 & 4096 & 27.92 & 37.15 & 40.34\\

\midrule
48.92 & 512 & 8 & 2048 & 26.61 & 35.46 & 40.47 \\
32.14 & 512 & 4 & 2048  & 28.31 &  38.53 & 42.87\\
23.75 & 512 & 2 & 2048 & 31.57 &  45.48 & 48.61\\
\midrule
16.07 & 256 & 8 & 1024 & 32.72   & 49.18 & 53.58\\
11.87 & 256 & 4 & 1024  & 34.23   & 53.18 & 56.26\\
9.77 & 256 & 2 & 1024  & 37.77   & 62.20 & 62.39\\
\midrule
5.94 & 128 & 8 & 512  & 41.38 & 70.05 & 72.53\\
4.89 & 128 & 4 & 512  & 43.69  & 77.73 & 77.38\\
4.36 & 128 & 2 & 512  & 47.37  & 87.18 & 84.05\\
\midrule
2.44 & 64 & 8 & 256  & 54.52   & 102.83 & 100.14 \\
2.18 & 64 & 4 & 256  & 57.86 & 113.23 & 105.72\\
2.05 & 64 & 2 & 256 & 62.78  & 124.61 & 114.69\\
\midrule
1.09 & 32 & 8 & 128 & 79.61 &  164.78 & 141.03\\
1.02 & 32 & 4 & 128  &  84.24 &  178.07 & 150.75\\
0.99 & 32 & 2 & 128 &  92.00 & 193.20 & 162.80\\ 
\bottomrule

\end{tabular*}
\caption{Parameter count and perplexity metrics across model configurations. ``PPL Simple (Simple Dataset)'' refers to the perplexity of simple models measured on a held-out filtered pre-training dataset. ``PPL Regular (Regular Dataset)'' refers to the perplexity of regular models measured on a held-out standard pre-training dataset. ``PPL Regular (Simple Dataset)'' refers to the perplexity of regular models measured on the held-out filtered pre-training dataset. As the model capacity increases, the ability to predict the evaluation data improves on both simple and regular models.}

% \footnotemark}
\label{tab:eval_loss_perplexity}
% \end{footnotesize}
\end{table*}
% \footnotetext{These metrics are reported on models trained with the pre-training dataset containing 2.1B tokens}

Table \ref{tab:piqa_arc_zero_shot} shows additional zero-shot accuracy results across all model configurations in both filtered and unfiltered datasets. On the PIQA task, we saw that simple models typically exhibit superior performance over regular models in a majority of configurations, regardless of whether the dataset was filtered or unfiltered. When focusing on the ARC-Easy filtered dataset, we found that simple models with larger hidden sizes (exceeding 64) consistently outperformed their regular counterparts on the filtered dataset. Conversely, with unfiltered ARC-Easy dataset, regular models demonstrated a higher performance level than the simple models.

Tables \ref{tab:blimp_comp_contrastive} and \ref{tab:blimp_unfilcomp_contrastive} show zero-shot accuracy results across various model configurations using the filtered and unfiltered datasets. Though simple models show a deterioration in performance compared to their results on the filtered dataset, we do see that the average scores tend to be better than the regular models on most configurations.

% \input{tables/table_few_shot_filtered}
% \input{tables/table_dataset_stats}
% PIQA / ARC zero-shot

\begin{table*}[htbp]
\begin{footnotesize}
\footnotesize
\begin{tabularx}{\textwidth}{XXXXXXXXXXXXXXX}
\toprule
Hidden Size & Num. Layers &                      PIQA (Filtered) &                    PIQA (Unfiltered) &                  ARC Easy (Filtered) &                ARC Easy (Unfiltered) \\
\midrule
       1024 &           8 &  0.653 $\textcolor{green}{\uparrow}$ &  0.642 $\textcolor{green}{\uparrow}$ &  0.398 $\textcolor{green}{\uparrow}$ &   0.330 $\textcolor{red}{\downarrow}$ \\
       1024 &           8 &                                0.565 &                                0.596 &                                0.372 &                                 0.370 \\
       1024 &           4 &  0.636 $\textcolor{green}{\uparrow}$ &  0.627 $\textcolor{green}{\uparrow}$ &  0.384 $\textcolor{green}{\uparrow}$ &   0.310 $\textcolor{red}{\downarrow}$ \\
       1024 &           4 &                                0.555 &                                0.576 &                                0.342 &                                0.353 \\
       1024 &           2 &   0.620 $\textcolor{green}{\uparrow}$ &  0.621 $\textcolor{green}{\uparrow}$ &  0.354 $\textcolor{green}{\uparrow}$ &  0.309 $\textcolor{red}{\downarrow}$ \\
       1024 &           2 &                                 0.560 &                                0.581 &                                0.342 &                                0.349 \\
        512 &           8 &   0.610 $\textcolor{green}{\uparrow}$ &  0.613 $\textcolor{green}{\uparrow}$ &  0.357 $\textcolor{green}{\uparrow}$ &  0.317 $\textcolor{red}{\downarrow}$ \\
        512 &           8 &                                0.556 &                                0.582 &                                0.292 &                                0.332 \\
        512 &           4 &  0.611 $\textcolor{green}{\uparrow}$ &   0.610 $\textcolor{green}{\uparrow}$ &  0.369 $\textcolor{green}{\uparrow}$ &  0.309 $\textcolor{red}{\downarrow}$ \\
        512 &           4 &                                0.524 &                                0.561 &                                0.345 &                                0.327 \\
        512 &           2 &  0.595 $\textcolor{green}{\uparrow}$ &  0.593 $\textcolor{green}{\uparrow}$ &                               0.319  &  0.303 $\textcolor{red}{\downarrow}$ \\
        512 &           2 &                                0.546 &                                0.569 &                                0.319 &                                0.325 \\
        256 &           8 &  0.587 $\textcolor{green}{\uparrow}$ &   0.580 $\textcolor{green}{\uparrow}$ &  0.351 $\textcolor{green}{\uparrow}$ &  0.309 $\textcolor{red}{\downarrow}$ \\
        256 &           8 &                                0.536 &                                0.564 &                                0.319 &                                0.312 \\
        256 &           4 &  0.585 $\textcolor{green}{\uparrow}$ &  0.584 $\textcolor{green}{\uparrow}$ &  0.325 $\textcolor{green}{\uparrow}$ &  0.298 $\textcolor{red}{\downarrow}$ \\
        256 &           4 &                                0.551 &                                0.568 &                                0.304 &                                0.315 \\
        256 &           2 &  0.589 $\textcolor{green}{\uparrow}$ &  0.594 $\textcolor{green}{\uparrow}$ &                               0.325  &   0.290 $\textcolor{red}{\downarrow}$ \\
        256 &           2 &                                0.553 &                                0.564 &                                0.325 &                                0.312 \\
        128 &           8 &   0.570 $\textcolor{green}{\uparrow}$ &  0.566 $\textcolor{green}{\uparrow}$ &  0.322 $\textcolor{green}{\uparrow}$ &  0.286 $\textcolor{red}{\downarrow}$ \\
        128 &           8 &                                0.528 &                                0.542 &                                0.286 &                                0.297 \\
        128 &           4 &  0.555 $\textcolor{green}{\uparrow}$ &  0.557 $\textcolor{green}{\uparrow}$ &   0.310 $\textcolor{green}{\uparrow}$ &  0.284 $\textcolor{red}{\downarrow}$ \\
        128 &           4 &                                0.532 &                                0.553 &                                0.292 &                                0.298 \\
        128 &           2 &  0.553 $\textcolor{green}{\uparrow}$ &   0.560 $\textcolor{green}{\uparrow}$ &  0.354 $\textcolor{green}{\uparrow}$ &  0.288 $\textcolor{red}{\downarrow}$ \\
        128 &           2 &                                0.523 &                                0.542 &                                0.298 &                                0.296 \\
         64 &           8 &  0.546 $\textcolor{green}{\uparrow}$ &  0.545 $\textcolor{green}{\uparrow}$ &  0.277 $\textcolor{red}{\downarrow}$ &  0.279 $\textcolor{red}{\downarrow}$ \\
         64 &           8 &                                0.519 &                                 0.540 &                                0.283 &                                0.284 \\
         64 &           4 &  0.543 $\textcolor{green}{\uparrow}$ &  0.547 $\textcolor{green}{\uparrow}$ &  0.271 $\textcolor{red}{\downarrow}$ &  0.282 $\textcolor{green}{\uparrow}$ \\
         64 &           4 &                                0.527 &                                0.535 &                                0.283 &                                0.277 \\
         64 &           2 &                               0.526  &  0.536 $\textcolor{red}{\downarrow}$ &  0.286 $\textcolor{red}{\downarrow}$ &  0.269 $\textcolor{red}{\downarrow}$ \\
         64 &           2 &                                0.526 &                                0.539 &                                0.298 &                                0.283 \\
         32 &           8 &  0.526 $\textcolor{green}{\uparrow}$ &  0.534 $\textcolor{green}{\uparrow}$ &                               0.292  &  0.265 $\textcolor{red}{\downarrow}$ \\
         32 &           8 &                                0.511 &                                0.533 &                                0.292 &                                0.275 \\
         32 &           4 &  0.526 $\textcolor{green}{\uparrow}$ &   0.530 $\textcolor{red}{\downarrow}$ &  0.271 $\textcolor{red}{\downarrow}$ &  0.264 $\textcolor{green}{\uparrow}$ \\
         32 &           4 &                                0.516 &                                0.533 &                                0.277 &                                0.263 \\
         32 &           2 &  0.541 $\textcolor{green}{\uparrow}$ &  0.532 $\textcolor{green}{\uparrow}$ &  0.254 $\textcolor{red}{\downarrow}$ &  0.264 $\textcolor{green}{\uparrow}$ \\
         32 &           2 &                                0.521 &                                0.529 &                                0.263 &                                0.259 \\
\bottomrule
\end{tabularx}
\end{footnotesize}
\caption{Zero-shot accuracy scores on PIQA and ARC Easy dataset. ``Filtered'' refers to the vocabulary filtered datasets and ``Unfiltered'' for standard datasets. For each hidden layer size and layer count, the table compares metrics for both simple and regular models.  The initial row for each hidden size and number of layers displays results from the simple model trained on simplified data, followed by a similar-sized regular model trained on regular data. Performance comparison is indicated by arrows next to the simple model's scores: a $\textcolor{green}{\uparrow}$ signifies the simple model outperforming the regular model, while a $\textcolor{red}{\downarrow}$ denotes the regular model performing better. An absence of arrows indicates comparable performance between the two models. We observe that on both the filtered and unfiltered datasets for the PIQA task, simple models generally outperform regular models in most configurations. Additionally, for the ARC Easy filtered dataset, larger simple models (with hidden sizes above 64) tend to surpass the performance of regular models. However, on the unfiltered ARC Easy dataset, it is observed that regular models outperform the simple ones.}
\label{tab:piqa_arc_zero_shot}
\end{table*}
% blimp filtered

\begin{table*}[h!]
\begin{tiny}
\begin{tabularx}{\textwidth}{XXXXXXXXXXXXXXX}
\toprule
Hidden & Num.&  Anaphor &  Argument &  Binding &  Control &  Determiner &  Ellipsis &  Filler &  Irregular &  Island &  NPI &  Quantifiers &  Subject &  Avg. \\
Size & Layers &  Agr. &  Str. &   &  /Raising &  Noun Agr. &   &  Gap &  Forms &  Effects &  Lic. &   &  Verb Agr. &  Score \\
\midrule
       1024 &           8 &  0.968 $\textcolor{green}{\uparrow}$ &  0.804 $\textcolor{green}{\uparrow}$ &  0.711 $\textcolor{red}{\downarrow}$ &  0.792 $\textcolor{green}{\uparrow}$ &  0.932 $\textcolor{green}{\uparrow}$ &  0.806 $\textcolor{green}{\uparrow}$ &  0.741 $\textcolor{green}{\uparrow}$ &  0.871 $\textcolor{green}{\uparrow}$ &  0.555 $\textcolor{red}{\downarrow}$ &  0.662 $\textcolor{green}{\uparrow}$ &  0.847 $\textcolor{green}{\uparrow}$ &  0.843 $\textcolor{green}{\uparrow}$ &  0.794 $\textcolor{green}{\uparrow}$ \\
       1024 &           8 &                                0.958 &                                0.783 &                                0.728 &                                0.775 &                                0.923 &                                0.722 &                                0.737 &                                0.801 &                                 0.610 &                                0.636 &                                0.713 &                                0.832 &                                0.768 \\
       1024 &           4 &  0.965 $\textcolor{green}{\uparrow}$ &  0.781 $\textcolor{green}{\uparrow}$ &  0.714 $\textcolor{red}{\downarrow}$ &  0.766 $\textcolor{green}{\uparrow}$ &  0.919 $\textcolor{red}{\downarrow}$ &  0.781 $\textcolor{green}{\uparrow}$ &  0.736 $\textcolor{green}{\uparrow}$ &   0.890 $\textcolor{green}{\uparrow}$ &  0.512 $\textcolor{red}{\downarrow}$ &  0.587 $\textcolor{green}{\uparrow}$ &    0.800 $\textcolor{green}{\uparrow}$ &  0.812 $\textcolor{green}{\uparrow}$ &  0.772 $\textcolor{green}{\uparrow}$ \\
       1024 &           4 &                                0.951 &                                0.777 &                                 0.730 &                                0.755 &                                0.924 &                                0.721 &                                0.732 &                                0.848 &                                0.573 &                                0.577 &                                0.593 &                                0.789 &                                0.748 \\
       1024 &           2 &  0.961 $\textcolor{green}{\uparrow}$ &  0.793 $\textcolor{green}{\uparrow}$ &  0.697 $\textcolor{green}{\uparrow}$ &  0.778 $\textcolor{green}{\uparrow}$ &  0.918 $\textcolor{green}{\uparrow}$ &  0.736 $\textcolor{green}{\uparrow}$ &  0.729 $\textcolor{green}{\uparrow}$ &  0.923 $\textcolor{green}{\uparrow}$ &  0.488 $\textcolor{red}{\downarrow}$ &  0.616 $\textcolor{red}{\downarrow}$ &  0.805 $\textcolor{green}{\uparrow}$ &  0.783 $\textcolor{green}{\uparrow}$ &  0.769 $\textcolor{green}{\uparrow}$ \\
       1024 &           2 &                                0.927 &                                0.755 &                                0.693 &                                0.729 &                                0.905 &                                0.689 &                                0.709 &                                0.873 &                                0.508 &                                0.624 &                                0.723 &                                0.744 &                                 0.740 \\
        512 &           8 &  0.938 $\textcolor{red}{\downarrow}$ &  0.785 $\textcolor{green}{\uparrow}$ &  0.688 $\textcolor{red}{\downarrow}$ &  0.762 $\textcolor{green}{\uparrow}$ &  0.909 $\textcolor{green}{\uparrow}$ &  0.761 $\textcolor{green}{\uparrow}$ &   0.720 $\textcolor{green}{\uparrow}$ &   0.950 $\textcolor{green}{\uparrow}$ &   0.560 $\textcolor{green}{\uparrow}$ &  0.608 $\textcolor{red}{\downarrow}$ &  0.707 $\textcolor{green}{\uparrow}$ &  0.816 $\textcolor{red}{\downarrow}$ &  0.767 $\textcolor{green}{\uparrow}$ \\
        512 &           8 &                                0.952 &                                0.756 &                                0.738 &                                0.745 &                                0.903 &                                 0.660 &                                0.698 &                                0.861 &                                0.537 &                                0.617 &                                0.662 &                                0.822 &                                0.746 \\
        512 &           4 &  0.961 $\textcolor{green}{\uparrow}$ &  0.787 $\textcolor{green}{\uparrow}$ &  0.681 $\textcolor{red}{\downarrow}$ &  0.779 $\textcolor{green}{\uparrow}$ &   0.920 $\textcolor{green}{\uparrow}$ &  0.757 $\textcolor{green}{\uparrow}$ &  0.728 $\textcolor{green}{\uparrow}$ &  0.931 $\textcolor{green}{\uparrow}$ &  0.536 $\textcolor{green}{\uparrow}$ &   0.570 $\textcolor{green}{\uparrow}$ &  0.778 $\textcolor{green}{\uparrow}$ &  0.827 $\textcolor{green}{\uparrow}$ &  0.771 $\textcolor{green}{\uparrow}$ \\
        512 &           4 &                                0.938 &                                0.753 &                                0.754 &                                0.748 &                                0.917 &                                0.654 &                                0.687 &                                0.851 &                                0.531 &                                0.557 &                                0.698 &                                0.813 &                                0.742 \\
        512 &           2 &  0.915 $\textcolor{green}{\uparrow}$ &  0.765 $\textcolor{green}{\uparrow}$ &  0.687 $\textcolor{red}{\downarrow}$ &  0.763 $\textcolor{green}{\uparrow}$ &  0.916 $\textcolor{green}{\uparrow}$ &  0.744 $\textcolor{green}{\uparrow}$ &  0.679 $\textcolor{red}{\downarrow}$ &  0.911 $\textcolor{green}{\uparrow}$ &  0.453 $\textcolor{red}{\downarrow}$ &  0.637 $\textcolor{green}{\uparrow}$ &  0.787 $\textcolor{green}{\uparrow}$ &  0.736 $\textcolor{green}{\uparrow}$ &  0.749 $\textcolor{green}{\uparrow}$ \\
        512 &           2 &                                 0.880 &                                0.727 &                                0.701 &                                0.736 &                                0.908 &                                0.679 &                                0.683 &                                0.896 &                                0.492 &                                0.464 &                                0.747 &                                0.733 &                                0.721 \\
        256 &           8 &  0.913 $\textcolor{green}{\uparrow}$ &  0.777 $\textcolor{green}{\uparrow}$ &  0.686 $\textcolor{red}{\downarrow}$ &  0.761 $\textcolor{green}{\uparrow}$ &  0.907 $\textcolor{green}{\uparrow}$ &  0.656 $\textcolor{green}{\uparrow}$ &  0.695 $\textcolor{green}{\uparrow}$ &  0.902 $\textcolor{green}{\uparrow}$ &  0.481 $\textcolor{red}{\downarrow}$ &  0.517 $\textcolor{red}{\downarrow}$ &  0.725 $\textcolor{green}{\uparrow}$ &  0.761 $\textcolor{green}{\uparrow}$ &  0.732 $\textcolor{green}{\uparrow}$ \\
        256 &           8 &                                0.897 &                                0.737 &                                0.733 &                                0.749 &                                0.878 &                                0.626 &                                0.663 &                                0.871 &                                0.516 &                                0.539 &                                0.718 &                                0.752 &                                0.723 \\
        256 &           4 &  0.936 $\textcolor{green}{\uparrow}$ &  0.778 $\textcolor{green}{\uparrow}$ &  0.691 $\textcolor{red}{\downarrow}$ &  0.772 $\textcolor{green}{\uparrow}$ &  0.893 $\textcolor{red}{\downarrow}$ &   0.670 $\textcolor{green}{\uparrow}$ &  0.669 $\textcolor{red}{\downarrow}$ &  0.893 $\textcolor{green}{\uparrow}$ &  0.506 $\textcolor{green}{\uparrow}$ &  0.623 $\textcolor{green}{\uparrow}$ &  0.713 $\textcolor{red}{\downarrow}$ &  0.778 $\textcolor{green}{\uparrow}$ &  0.744 $\textcolor{green}{\uparrow}$ \\
        256 &           4 &                                0.867 &                                0.744 &                                0.716 &                                0.715 &                                0.898 &                                0.617 &                                0.676 &                                0.837 &                                0.505 &                                 0.560 &                                0.736 &                                0.714 &                                0.715 \\
        256 &           2 &  0.893 $\textcolor{green}{\uparrow}$ &  0.744 $\textcolor{green}{\uparrow}$ &   0.690 $\textcolor{red}{\downarrow}$ &  0.749 $\textcolor{green}{\uparrow}$ &   0.890 $\textcolor{green}{\uparrow}$ &  0.674 $\textcolor{green}{\uparrow}$ &  0.678 $\textcolor{green}{\uparrow}$ &  0.918 $\textcolor{green}{\uparrow}$ &  0.451 $\textcolor{red}{\downarrow}$ &  0.556 $\textcolor{red}{\downarrow}$ &  0.742 $\textcolor{green}{\uparrow}$ &  0.707 $\textcolor{green}{\uparrow}$ &  0.724 $\textcolor{green}{\uparrow}$ \\
        256 &           2 &                                0.798 &                                0.715 &                                0.691 &                                0.718 &                                0.852 &                                0.605 &                                 0.660 &                                0.913 &                                0.481 &                                0.575 &                                 0.720 &                                0.639 &                                0.697 \\
        128 &           8 &  0.884 $\textcolor{green}{\uparrow}$ &  0.734 $\textcolor{green}{\uparrow}$ &   0.720 $\textcolor{green}{\uparrow}$ &  0.761 $\textcolor{green}{\uparrow}$ &  0.895 $\textcolor{green}{\uparrow}$ &  0.654 $\textcolor{green}{\uparrow}$ &  0.662 $\textcolor{red}{\downarrow}$ &  0.935 $\textcolor{green}{\uparrow}$ &   0.450 $\textcolor{red}{\downarrow}$ &  0.623 $\textcolor{green}{\uparrow}$ &  0.714 $\textcolor{green}{\uparrow}$ &  0.711 $\textcolor{green}{\uparrow}$ &  0.729 $\textcolor{green}{\uparrow}$ \\
        128 &           8 &                                0.785 &                                0.694 &                                0.698 &                                0.731 &                                0.859 &                                0.591 &                                0.668 &                                0.909 &                                0.516 &                                0.423 &                                0.697 &                                0.649 &                                0.685 \\
        128 &           4 &  0.854 $\textcolor{green}{\uparrow}$ &  0.754 $\textcolor{green}{\uparrow}$ &  0.689 $\textcolor{red}{\downarrow}$ &  0.754 $\textcolor{green}{\uparrow}$ &  0.885 $\textcolor{green}{\uparrow}$ &  0.624 $\textcolor{red}{\downarrow}$ &  0.679 $\textcolor{green}{\uparrow}$ &  0.882 $\textcolor{red}{\downarrow}$ &  0.454 $\textcolor{red}{\downarrow}$ &   0.550 $\textcolor{red}{\downarrow}$ &  0.726 $\textcolor{red}{\downarrow}$ &  0.754 $\textcolor{green}{\uparrow}$ &  0.717 $\textcolor{green}{\uparrow}$ \\
        128 &           4 &                                0.774 &                                0.699 &                                  0.700 &                                0.703 &                                0.838 &                                0.632 &                                 0.660 &                                0.886 &                                0.484 &                                0.599 &                                0.747 &                                0.642 &                                0.697 \\
        128 &           2 &  0.847 $\textcolor{green}{\uparrow}$ &  0.727 $\textcolor{green}{\uparrow}$ &  0.681 $\textcolor{green}{\uparrow}$ &  0.756 $\textcolor{green}{\uparrow}$ &  0.864 $\textcolor{green}{\uparrow}$ &  0.675 $\textcolor{green}{\uparrow}$ &  0.667 $\textcolor{green}{\uparrow}$ &  0.908 $\textcolor{green}{\uparrow}$ &  0.415 $\textcolor{red}{\downarrow}$ &  0.631 $\textcolor{green}{\uparrow}$ &  0.683 $\textcolor{green}{\uparrow}$ &  0.677 $\textcolor{green}{\uparrow}$ &  0.711 $\textcolor{green}{\uparrow}$ \\
        128 &           2 &                                0.802 &                                0.667 &                                0.662 &                                0.673 &                                0.842 &                                0.593 &                                0.626 &                                0.821 &                                0.481 &                                0.586 &                                0.654 &                                0.627 &                                0.669 \\
         64 &           8 &  0.847 $\textcolor{green}{\uparrow}$ &   0.720 $\textcolor{green}{\uparrow}$ &  0.674 $\textcolor{green}{\uparrow}$ &  0.692 $\textcolor{green}{\uparrow}$ &  0.873 $\textcolor{green}{\uparrow}$ &  0.575 $\textcolor{green}{\uparrow}$ &  0.647 $\textcolor{green}{\uparrow}$ &   0.890 $\textcolor{green}{\uparrow}$ &  0.485 $\textcolor{green}{\uparrow}$ &  0.605 $\textcolor{green}{\uparrow}$ &  0.803 $\textcolor{green}{\uparrow}$ &  0.672 $\textcolor{green}{\uparrow}$ &  0.707 $\textcolor{green}{\uparrow}$ \\
         64 &           8 &                                0.696 &                                0.673 &                                 0.640 &                                 0.690 &                                 0.830 &                                0.543 &                                0.624 &                                0.866 &                                0.472 &                                0.479 &                                0.639 &                                0.606 &                                0.646 \\
         64 &           4 &  0.807 $\textcolor{green}{\uparrow}$ &  0.707 $\textcolor{green}{\uparrow}$ &  0.669 $\textcolor{red}{\downarrow}$ &  0.699 $\textcolor{green}{\uparrow}$ &  0.877 $\textcolor{green}{\uparrow}$ &  0.556 $\textcolor{red}{\downarrow}$ &  0.662 $\textcolor{green}{\uparrow}$ &  0.867 $\textcolor{green}{\uparrow}$ &  0.455 $\textcolor{green}{\uparrow}$ &  0.594 $\textcolor{green}{\uparrow}$ &  0.572 $\textcolor{red}{\downarrow}$ &   0.610 $\textcolor{red}{\downarrow}$ &  0.673 $\textcolor{green}{\uparrow}$ \\
         64 &           4 &                                0.697 &                                0.671 &                                 0.670 &                                0.656 &                                0.833 &                                0.588 &                                0.596 &                                0.851 &                                0.428 &                                0.565 &                                0.589 &                                0.613 &                                0.646 \\
         64 &           2 &   0.790 $\textcolor{green}{\uparrow}$ &  0.684 $\textcolor{green}{\uparrow}$ &  0.659 $\textcolor{green}{\uparrow}$ &  0.683 $\textcolor{green}{\uparrow}$ &  0.874 $\textcolor{green}{\uparrow}$ &  0.577 $\textcolor{red}{\downarrow}$ &   0.660 $\textcolor{green}{\uparrow}$ &  0.847 $\textcolor{green}{\uparrow}$ &  0.408 $\textcolor{red}{\downarrow}$ &  0.527 $\textcolor{red}{\downarrow}$ &  0.665 $\textcolor{green}{\uparrow}$ &  0.629 $\textcolor{green}{\uparrow}$ &  0.667 $\textcolor{green}{\uparrow}$ \\
         64 &           2 &                                0.637 &                                0.646 &                                0.647 &                                0.623 &                                0.794 &                                0.596 &                                0.635 &                                0.826 &                                0.409 &                                0.583 &                                0.567 &                                 0.570 &                                0.628 \\
         32 &           8 &  0.746 $\textcolor{red}{\downarrow}$ &  0.674 $\textcolor{green}{\uparrow}$ &   0.650 $\textcolor{green}{\uparrow}$ &  0.632 $\textcolor{green}{\uparrow}$ &  0.809 $\textcolor{green}{\uparrow}$ &  0.498 $\textcolor{red}{\downarrow}$ &  0.631 $\textcolor{green}{\uparrow}$ &  0.837 $\textcolor{red}{\downarrow}$ &  0.482 $\textcolor{green}{\uparrow}$ &    0.500 $\textcolor{green}{\uparrow}$ &  0.659 $\textcolor{green}{\uparrow}$ &  0.574 $\textcolor{red}{\downarrow}$ &  0.641 $\textcolor{green}{\uparrow}$ \\
         32 &           8 &                                0.762 &                                0.649 &                                0.615 &                                0.589 &                                0.786 &                                0.521 &                                0.612 &                                0.839 &                                0.475 &                                0.498 &                                0.604 &                                0.593 &                                0.629 \\
         32 &           4 &  0.479 $\textcolor{red}{\downarrow}$ &  0.619 $\textcolor{red}{\downarrow}$ &  0.643 $\textcolor{green}{\uparrow}$ &  0.644 $\textcolor{green}{\uparrow}$ &  0.741 $\textcolor{green}{\uparrow}$ &  0.489 $\textcolor{green}{\uparrow}$ &  0.621 $\textcolor{green}{\uparrow}$ &  0.782 $\textcolor{red}{\downarrow}$ &  0.519 $\textcolor{green}{\uparrow}$ &  0.505 $\textcolor{red}{\downarrow}$ &  0.559 $\textcolor{red}{\downarrow}$ &  0.592 $\textcolor{green}{\uparrow}$ &  0.599 $\textcolor{red}{\downarrow}$ \\
         32 &           4 &                                0.612 &                                0.656 &                                0.623 &                                0.609 &                                0.731 &                                0.488 &                                0.607 &                                0.797 &                                0.453 &                                0.584 &                                 0.660 &                                0.544 &                                0.614 \\
         32 &           2 &  0.723 $\textcolor{green}{\uparrow}$ &   0.640 $\textcolor{green}{\uparrow}$ &  0.635 $\textcolor{green}{\uparrow}$ &  0.631 $\textcolor{green}{\uparrow}$ &   0.740 $\textcolor{green}{\uparrow}$ &  0.485 $\textcolor{green}{\uparrow}$ &  0.641 $\textcolor{green}{\uparrow}$ &  0.855 $\textcolor{green}{\uparrow}$ &  0.473 $\textcolor{red}{\downarrow}$ &  0.618 $\textcolor{green}{\uparrow}$ &  0.654 $\textcolor{green}{\uparrow}$ &  0.587 $\textcolor{green}{\uparrow}$ &   0.640 $\textcolor{green}{\uparrow}$ \\
         32 &           2 &                                0.591 &                                0.638 &                                0.615 &                                0.626 &                                0.686 &                                 0.420 &                                0.603 &                                0.733 &                                0.489 &                                0.606 &                                0.476 &                                0.526 &                                0.584 \\
\bottomrule
\end{tabularx}
\end{tiny}
\caption{Zero-shot accuracy scores on the vocabulary filtered datasets in the BLiMP benchmark. For each hidden layer size and layer count, the table compares metrics for both simple and regular models.  The initial row for each hidden size and number of layers displays results from the simple model trained on simplified data, followed by a similar-sized regular model trained on regular data. Performance comparison is indicated by arrows next to the simple model's scores: a $\textcolor{green}{\uparrow}$ signifies the simple model outperforming the regular model, while a $\textcolor{red}{\downarrow}$ denotes the regular model performing better. An absence of arrows indicates comparable performance between the two models. We find that the average score of simple models tends to surpass regular models in most configurations.}
\label{tab:blimp_comp_contrastive}
\end{table*}
\begin{table*}[h!]
\begin{tiny}
\begin{tabularx}{\textwidth}{XXXXXXXXXXXXXXX}
\toprule
Hidden & Num.&  Anaphor &  Argument &  Binding &  Control &  Determiner &  Ellipsis &  Filler &  Irregular &  Island &  NPI &  Quantifiers &  Subject &  Avg. \\
Size & Layers &  Agr. &  Str. &   &  /Raising &  Noun Agr. &   &  Gap &  Forms &  Effects &  Lic. &   &  Verb Agr. &  Score \\
\midrule
       1024 &           8 &  0.919 $\textcolor{red}{\downarrow}$ &  0.776 $\textcolor{red}{\downarrow}$ &  0.719 $\textcolor{red}{\downarrow}$ &  0.757 $\textcolor{red}{\downarrow}$ &  0.891 $\textcolor{red}{\downarrow}$ &  0.782 $\textcolor{green}{\uparrow}$ &  0.725 $\textcolor{red}{\downarrow}$ &  0.883 $\textcolor{green}{\uparrow}$ &  0.563 $\textcolor{red}{\downarrow}$ &  0.644 $\textcolor{green}{\uparrow}$ &  0.846 $\textcolor{green}{\uparrow}$ &  0.786 $\textcolor{red}{\downarrow}$ &  0.774 $\textcolor{green}{\uparrow}$ \\
       1024 &           8 &                                0.972 &                                 0.780 &                                0.728 &                                0.761 &                                0.917 &                                0.735 &                                0.728 &                                0.845 &                                0.607 &                                 0.630 &                                0.719 &                                 0.830 &                                0.771 \\
       1024 &           4 &  0.918 $\textcolor{red}{\downarrow}$ &                               0.749  &  0.709 $\textcolor{red}{\downarrow}$ &  0.742 $\textcolor{green}{\uparrow}$ &   0.870 $\textcolor{red}{\downarrow}$ &  0.753 $\textcolor{green}{\uparrow}$ &  0.726 $\textcolor{green}{\uparrow}$ &  0.885 $\textcolor{green}{\uparrow}$ &  0.502 $\textcolor{red}{\downarrow}$ &                               0.568  &  0.797 $\textcolor{green}{\uparrow}$ &  0.762 $\textcolor{red}{\downarrow}$ &                               0.748  \\
       1024 &           4 &                                0.955 &                                0.749 &                                0.749 &                                0.737 &                                0.916 &                                0.735 &                                0.724 &                                0.877 &                                0.578 &                                0.568 &                                  0.600 &                                0.783 &                                0.748 \\
       1024 &           2 &  0.864 $\textcolor{red}{\downarrow}$ &  0.744 $\textcolor{red}{\downarrow}$ &  0.705 $\textcolor{red}{\downarrow}$ &  0.735 $\textcolor{green}{\uparrow}$ &  0.865 $\textcolor{red}{\downarrow}$ &  0.706 $\textcolor{green}{\uparrow}$ &  0.717 $\textcolor{green}{\uparrow}$ &  0.908 $\textcolor{green}{\uparrow}$ &  0.496 $\textcolor{red}{\downarrow}$ &  0.606 $\textcolor{red}{\downarrow}$ &  0.798 $\textcolor{green}{\uparrow}$ &  0.725 $\textcolor{green}{\uparrow}$ &  0.739 $\textcolor{green}{\uparrow}$ \\
       1024 &           2 &                                0.903 &                                0.745 &                                0.716 &                                0.716 &                                0.886 &                                 0.690 &                                0.702 &                                0.894 &                                0.506 &                                0.621 &                                0.721 &                                0.721 &                                0.735 \\
        512 &           8 &  0.912 $\textcolor{red}{\downarrow}$ &   0.740 $\textcolor{red}{\downarrow}$ &  0.696 $\textcolor{red}{\downarrow}$ &  0.734 $\textcolor{green}{\uparrow}$ &  0.849 $\textcolor{red}{\downarrow}$ &  0.723 $\textcolor{green}{\uparrow}$ &  0.707 $\textcolor{green}{\uparrow}$ &  0.924 $\textcolor{green}{\uparrow}$ &  0.561 $\textcolor{green}{\uparrow}$ &  0.596 $\textcolor{red}{\downarrow}$ &  0.708 $\textcolor{green}{\uparrow}$ &  0.768 $\textcolor{red}{\downarrow}$ &  0.743 $\textcolor{red}{\downarrow}$ \\
        512 &           8 &                                0.963 &                                0.762 &                                0.731 &                                0.725 &                                0.907 &                                0.681 &                                0.695 &                                0.877 &                                 0.540 &                                 0.620 &                                0.671 &                                0.805 &                                0.748 \\
        512 &           4 &  0.902 $\textcolor{red}{\downarrow}$ &  0.735 $\textcolor{red}{\downarrow}$ &  0.677 $\textcolor{red}{\downarrow}$ &  0.744 $\textcolor{green}{\uparrow}$ &  0.869 $\textcolor{red}{\downarrow}$ &   0.710 $\textcolor{green}{\uparrow}$ &   0.720 $\textcolor{green}{\uparrow}$ &  0.912 $\textcolor{green}{\uparrow}$ &  0.537 $\textcolor{green}{\uparrow}$ &  0.566 $\textcolor{green}{\uparrow}$ &  0.755 $\textcolor{green}{\uparrow}$ &  0.769 $\textcolor{red}{\downarrow}$ &  0.741 $\textcolor{green}{\uparrow}$ \\
        512 &           4 &                                0.942 &                                0.757 &                                0.733 &                                0.743 &                                0.885 &                                0.652 &                                0.684 &                                0.892 &                                0.528 &                                0.551 &                                0.704 &                                  0.800 &                                0.739 \\
        512 &           2 &  0.855 $\textcolor{red}{\downarrow}$ &  0.714 $\textcolor{red}{\downarrow}$ &   0.690 $\textcolor{red}{\downarrow}$ &  0.727 $\textcolor{green}{\uparrow}$ &  0.862 $\textcolor{red}{\downarrow}$ &  0.685 $\textcolor{green}{\uparrow}$ &  0.674 $\textcolor{red}{\downarrow}$ &  0.887 $\textcolor{green}{\uparrow}$ &  0.443 $\textcolor{red}{\downarrow}$ &  0.628 $\textcolor{green}{\uparrow}$ &  0.787 $\textcolor{green}{\uparrow}$ &  0.674 $\textcolor{red}{\downarrow}$ &  0.719 $\textcolor{green}{\uparrow}$ \\
        512 &           2 &                                0.856 &                                0.725 &                                0.715 &                                0.724 &                                0.881 &                                0.681 &                                0.676 &                                0.883 &                                0.488 &                                0.473 &                                0.757 &                                0.709 &                                0.714 \\
        256 &           8 &  0.856 $\textcolor{red}{\downarrow}$ &  0.725 $\textcolor{red}{\downarrow}$ &  0.697 $\textcolor{red}{\downarrow}$ &                               0.728  &  0.863 $\textcolor{green}{\uparrow}$ &  0.625 $\textcolor{green}{\uparrow}$ &   0.690 $\textcolor{green}{\uparrow}$ &  0.874 $\textcolor{green}{\uparrow}$ &  0.495 $\textcolor{red}{\downarrow}$ &  0.509 $\textcolor{red}{\downarrow}$ &  0.729 $\textcolor{green}{\uparrow}$ &  0.704 $\textcolor{red}{\downarrow}$ &  0.708 $\textcolor{red}{\downarrow}$ \\
        256 &           8 &                                0.899 &                                0.733 &                                0.751 &                                0.728 &                                0.856 &                                0.621 &                                 0.670 &                                0.873 &                                0.518 &                                0.536 &                                0.722 &                                0.747 &                                0.721 \\
        256 &           4 &  0.815 $\textcolor{red}{\downarrow}$ &  0.724 $\textcolor{red}{\downarrow}$ &  0.713 $\textcolor{red}{\downarrow}$ &  0.731 $\textcolor{green}{\uparrow}$ &  0.838 $\textcolor{red}{\downarrow}$ &   0.650 $\textcolor{green}{\uparrow}$ &  0.681 $\textcolor{green}{\uparrow}$ &  0.891 $\textcolor{green}{\uparrow}$ &  0.509 $\textcolor{red}{\downarrow}$ &  0.616 $\textcolor{green}{\uparrow}$ &  0.705 $\textcolor{red}{\downarrow}$ &  0.712 $\textcolor{green}{\uparrow}$ &  0.715 $\textcolor{green}{\uparrow}$ \\
        256 &           4 &                                 0.870 &                                0.729 &                                0.728 &                                  0.700 &                                0.864 &                                0.614 &                                0.679 &                                0.853 &                                0.515 &                                0.547 &                                 0.740 &                                0.695 &                                0.711 \\
        256 &           2 &  0.818 $\textcolor{green}{\uparrow}$ &  0.689 $\textcolor{red}{\downarrow}$ &  0.706 $\textcolor{red}{\downarrow}$ &  0.711 $\textcolor{green}{\uparrow}$ &  0.821 $\textcolor{green}{\uparrow}$ &  0.636 $\textcolor{green}{\uparrow}$ &  0.675 $\textcolor{green}{\uparrow}$ &  0.892 $\textcolor{red}{\downarrow}$ &  0.441 $\textcolor{red}{\downarrow}$ &  0.552 $\textcolor{red}{\downarrow}$ &  0.737 $\textcolor{green}{\uparrow}$ &  0.656 $\textcolor{green}{\uparrow}$ &  0.695 $\textcolor{green}{\uparrow}$ \\
        256 &           2 &                                0.743 &                                  0.700 &                                0.707 &                                0.696 &                                0.812 &                                0.627 &                                0.665 &                                0.894 &                                0.486 &                                0.559 &                                0.732 &                                0.627 &                                0.687 \\
        128 &           8 &  0.772 $\textcolor{green}{\uparrow}$ &  0.674 $\textcolor{red}{\downarrow}$ &  0.728 $\textcolor{green}{\uparrow}$ &  0.718 $\textcolor{green}{\uparrow}$ &  0.823 $\textcolor{red}{\downarrow}$ &  0.593 $\textcolor{green}{\uparrow}$ &  0.665 $\textcolor{red}{\downarrow}$ &  0.884 $\textcolor{red}{\downarrow}$ &  0.452 $\textcolor{red}{\downarrow}$ &  0.606 $\textcolor{green}{\uparrow}$ &  0.727 $\textcolor{green}{\uparrow}$ &  0.662 $\textcolor{green}{\uparrow}$ &  0.692 $\textcolor{green}{\uparrow}$ \\
        128 &           8 &                                 0.760 &                                0.696 &                                0.717 &                                0.702 &                                0.829 &                                0.584 &                                0.669 &                                0.893 &                                0.517 &                                0.422 &                                0.709 &                                0.634 &                                0.678 \\
        128 &           4 &  0.751 $\textcolor{green}{\uparrow}$ &  0.685 $\textcolor{red}{\downarrow}$ &  0.698 $\textcolor{red}{\downarrow}$ &  0.706 $\textcolor{green}{\uparrow}$ &   0.830 $\textcolor{green}{\uparrow}$ &  0.606 $\textcolor{green}{\uparrow}$ &  0.678 $\textcolor{green}{\uparrow}$ &  0.859 $\textcolor{red}{\downarrow}$ &  0.458 $\textcolor{red}{\downarrow}$ &  0.544 $\textcolor{red}{\downarrow}$ &   0.720 $\textcolor{red}{\downarrow}$ &  0.692 $\textcolor{green}{\uparrow}$ &  0.686 $\textcolor{green}{\uparrow}$ \\
        128 &           4 &                                0.695 &                                 0.69 &                                0.727 &                                0.686 &                                0.807 &                                0.594 &                                0.664 &                                0.881 &                                0.484 &                                0.588 &                                0.751 &                                0.622 &                                0.682 \\
        128 &           2 &  0.752 $\textcolor{green}{\uparrow}$ &  0.659 $\textcolor{red}{\downarrow}$ &  0.696 $\textcolor{green}{\uparrow}$ &  0.707 $\textcolor{green}{\uparrow}$ &  0.795 $\textcolor{green}{\uparrow}$ &  0.608 $\textcolor{green}{\uparrow}$ &   0.660 $\textcolor{green}{\uparrow}$ &  0.875 $\textcolor{green}{\uparrow}$ &  0.432 $\textcolor{red}{\downarrow}$ &  0.617 $\textcolor{green}{\uparrow}$ &  0.696 $\textcolor{green}{\uparrow}$ &  0.609 $\textcolor{green}{\uparrow}$ &  0.676 $\textcolor{green}{\uparrow}$ \\
        128 &           2 &                                0.738 &                                0.668 &                                0.676 &                                0.663 &                                0.786 &                                0.562 &                                0.645 &                                0.833 &                                0.476 &                                0.582 &                                0.667 &                                0.594 &                                0.658 \\
         64 &           8 &  0.692 $\textcolor{green}{\uparrow}$ &   0.650 $\textcolor{red}{\downarrow}$ &  0.699 $\textcolor{green}{\uparrow}$ &  0.675 $\textcolor{green}{\uparrow}$ &  0.808 $\textcolor{green}{\uparrow}$ &  0.556 $\textcolor{green}{\uparrow}$ &  0.663 $\textcolor{green}{\uparrow}$ &  0.867 $\textcolor{green}{\uparrow}$ &  0.493 $\textcolor{green}{\uparrow}$ &  0.593 $\textcolor{green}{\uparrow}$ &  0.794 $\textcolor{green}{\uparrow}$ &  0.607 $\textcolor{green}{\uparrow}$ &  0.675 $\textcolor{green}{\uparrow}$ \\
         64 &           8 &                                0.597 &                                 0.670 &                                0.664 &                                0.658 &                                0.781 &                                0.528 &                                0.642 &                                0.859 &                                0.465 &                                 0.470 &                                0.645 &                                0.586 &                                 0.630 \\
         64 &           4 &  0.638 $\textcolor{green}{\uparrow}$ &  0.642 $\textcolor{red}{\downarrow}$ &   0.680 $\textcolor{red}{\downarrow}$ &  0.686 $\textcolor{green}{\uparrow}$ &  0.807 $\textcolor{green}{\uparrow}$ &  0.506 $\textcolor{red}{\downarrow}$ &  0.665 $\textcolor{green}{\uparrow}$ &  0.845 $\textcolor{green}{\uparrow}$ &  0.446 $\textcolor{green}{\uparrow}$ &  0.578 $\textcolor{green}{\uparrow}$ &  0.575 $\textcolor{red}{\downarrow}$ &  0.567 $\textcolor{red}{\downarrow}$ &  0.636 $\textcolor{green}{\uparrow}$ \\
         64 &           4 &                                0.569 &                                0.657 &                                0.692 &                                0.651 &                                0.783 &                                0.541 &                                 0.620 &                                0.829 &                                 0.420 &                                0.557 &                                0.598 &                                0.581 &                                0.625 \\
         64 &           2 &  0.634 $\textcolor{green}{\uparrow}$ &  0.626 $\textcolor{red}{\downarrow}$ &  0.678 $\textcolor{green}{\uparrow}$ &  0.668 $\textcolor{green}{\uparrow}$ &  0.809 $\textcolor{green}{\uparrow}$ &  0.536 $\textcolor{red}{\downarrow}$ &  0.661 $\textcolor{green}{\uparrow}$ &  0.831 $\textcolor{green}{\uparrow}$ &  0.408 $\textcolor{green}{\uparrow}$ &  0.519 $\textcolor{red}{\downarrow}$ &  0.664 $\textcolor{green}{\uparrow}$ &  0.591 $\textcolor{green}{\uparrow}$ &  0.635 $\textcolor{green}{\uparrow}$ \\
         64 &           2 &                                0.585 &                                0.637 &                                0.659 &                                 0.620 &                                0.752 &                                 0.590 &                                0.638 &                                0.816 &                                  0.400 &                                0.585 &                                0.565 &                                0.551 &                                0.617 \\
         32 &           8 &   0.740 $\textcolor{green}{\uparrow}$ &  0.615 $\textcolor{red}{\downarrow}$ &  0.672 $\textcolor{green}{\uparrow}$ &  0.619 $\textcolor{green}{\uparrow}$ &  0.744 $\textcolor{green}{\uparrow}$ &  0.461 $\textcolor{red}{\downarrow}$ &  0.635 $\textcolor{green}{\uparrow}$ &  0.818 $\textcolor{red}{\downarrow}$ &  0.465 $\textcolor{red}{\downarrow}$ &  0.499 $\textcolor{green}{\uparrow}$ &   0.660 $\textcolor{green}{\uparrow}$ &  0.536 $\textcolor{red}{\downarrow}$ &  0.622 $\textcolor{green}{\uparrow}$ \\
         32 &           8 &                                0.513 &                                 0.630 &                                0.641 &                                0.592 &                                0.709 &                                0.498 &                                0.611 &                                0.831 &                                0.469 &                                0.496 &                                0.602 &                                0.557 &                                0.596 \\
         32 &           4 &  0.392 $\textcolor{red}{\downarrow}$ &  0.575 $\textcolor{red}{\downarrow}$ &  0.655 $\textcolor{green}{\uparrow}$ &  0.613 $\textcolor{green}{\uparrow}$ &  0.707 $\textcolor{green}{\uparrow}$ &  0.461 $\textcolor{red}{\downarrow}$ &  0.625 $\textcolor{green}{\uparrow}$ &  0.767 $\textcolor{red}{\downarrow}$ &  0.494 $\textcolor{green}{\uparrow}$ &    0.500 $\textcolor{red}{\downarrow}$ &  0.565 $\textcolor{red}{\downarrow}$ &  0.543 $\textcolor{green}{\uparrow}$ &  0.575 $\textcolor{red}{\downarrow}$ \\
         32 &           4 &                                0.539 &                                0.635 &                                0.653 &                                0.606 &                                0.663 &                                0.477 &                                0.618 &                                  0.800 &                                0.451 &                                0.577 &                                0.659 &                                0.535 &                                0.601 \\
         32 &           2 &  0.567 $\textcolor{green}{\uparrow}$ &  0.587 $\textcolor{red}{\downarrow}$ &  0.663 $\textcolor{green}{\uparrow}$ &  0.614 $\textcolor{green}{\uparrow}$ &  0.688 $\textcolor{green}{\uparrow}$ &  0.432 $\textcolor{green}{\uparrow}$ &  0.646 $\textcolor{green}{\uparrow}$ &  0.831 $\textcolor{green}{\uparrow}$ &  0.468 $\textcolor{red}{\downarrow}$ &  0.621 $\textcolor{green}{\uparrow}$ &  0.647 $\textcolor{green}{\uparrow}$ &  0.533 $\textcolor{green}{\uparrow}$ &  0.608 $\textcolor{green}{\uparrow}$ \\
         32 &           2 &                                0.499 &                                0.616 &                                0.633 &                                0.599 &                                0.625 &                                0.424 &                                0.609 &                                0.759 &                                0.486 &                                0.598 &                                 0.480 &                                0.513 &                                 0.570 \\
\bottomrule
\end{tabularx}
\end{tiny}
\caption{Zero-shot accuracy scores on the datasets in the BLiMP benchmark. For each hidden layer size and layer count, the table compares metrics for both simple and regular models.  The initial row for each hidden size and number of layers displays results from the simple model trained on simplified data, followed by a similar-sized regular model trained on regular data. Performance comparison is indicated by arrows next to the simple model's scores: a $\textcolor{green}{\uparrow}$ signifies the simple model outperforming the regular model, while a $\textcolor{red}{\downarrow}$ denotes the regular model performing better. An absence of arrows indicates comparable performance between the two models. We find that the average score of simple models tends to surpass regular models in most configurations.}
\label{tab:blimp_unfilcomp_contrastive}
\end{table*}

% \section{Statistics of the Evaluation Datasets}
% \label{sec:appendix_stats}

\section{Rotary Position Embeddings and Position Interpolation}
\label{sec:appendix_rope}

For models pre-trained on 2.1B tokens we use context length of 128. However, datasets such as PIQA and ARC-Easy contain examples that span more than the pre-trained context length. To extend context window sizes beyond 128, we use Position Interpolation \citep{chen2023extending} on PIQA and ARC-Easy datasets. We use a scaling factor of 8 which allows to have context window of 1024. 

Based on the findings presented in a study conducted by \citet{liu2023scaling}, we conducted an exploratory experiment for deciding the base value for the rotary positions embeddings \citep{su2023roformer}. For the pre-training sequence length of 128, we observed better length extension results (with PI \citep{chen2023extending}) for the base value of 20, compared to the widely used 10,000. Our results were in agreement with the findings presented by \citet{liu2023scaling}. Hence, we used a base value of 20 for pre-training language models.

In our downstream evaluation, we utilized PI for context length extension only for the PIQA and ARC-Easy datasets. 
% According to the length distribution in the XXX dataset (XXX is the 90th percentile), 
We used a scale of 8 for extending the pre-training context length to 1024. With the PI scale of 8, we evaluated the model on the government report dataset \citep{huang-etal-2021-efficient} and observed a decreasing perplexity from a context length of 64 to 1,024 in our exploratory experiment. 

\end{document}